\documentclass{article}

\usepackage[english]{babel}

\usepackage[letterpaper,top=2cm,bottom=2cm,left=3cm,right=3cm,marginparwidth=1.75cm]{geometry}

\usepackage{amsmath}
\usepackage{amssymb}
\usepackage{graphicx}
\usepackage{mathrsfs}
\usepackage{booktabs}
\usepackage{changepage}
\usepackage{bm}
\usepackage{authblk}
\usepackage[colorlinks=true, allcolors=blue]{hyperref}

\title{SynerMix: Synergistic Mixup Solution for Enhanced Intra-Class Cohesion and Inter-Class Separability in Image Classification}
\author{Ye Xu\textsuperscript{1}\footnote{Corresponding author}, Ya Gao\textsuperscript{1}, Xiaorong Qiu\textsuperscript{1}, Yang Chen\textsuperscript{1}, Ying Ji\textsuperscript{1}}

\affil{\textsuperscript{1}School of IoT Engineering, Wuxi Institute of Technology\\}

\begin{document}
\maketitle

\begin{abstract}
To address the issues of MixUp and its variants (e.g., Manifold MixUp) in image classification tasks—namely, their neglect of mixing within the same class (intra-class mixup) and their inadequacy in enhancing intra-class cohesion through their mixing operations—we propose a novel mixup method named \emph{SynerMix-Intra} and, building upon this, introduce a synergistic mixup solution named \emph{SynerMix}. \emph{SynerMix-Intra} specifically targets intra-class mixup to bolster intra-class cohesion, a feature not addressed by current mixup methods. For each mini-batch, it leverages feature representations of unaugmented original images from each class to generate a synthesized feature representation through random linear interpolation. All synthesized representations are then fed into the classification and loss layers to calculate an average classification loss that significantly enhances intra-class cohesion. Furthermore, \emph{SynerMix} combines \emph{SynerMix-Intra} with an existing mixup approach (e.g., MixUp, Manifold MixUp), which primarily focuses on inter-class mixup and has the benefit of enhancing inter-class separability. In doing so, it integrates both inter- and intra-class mixup in a balanced way while concurrently improving intra-class cohesion and inter-class separability. Experimental results on six datasets show that \emph{SynerMix} achieves a 0.1\% to 3.43\% higher accuracy than the best of either MixUp or \emph{SynerMix-Intra} alone, averaging a 1.16\% gain. It also surpasses the top-performer of either Manifold MixUp or \emph{SynerMix-Intra} by 0.12\% to 5.16\%, with an average gain of 1.11\%. Given that \emph{SynerMix} is model-agnostic, it holds significant potential for application in other domains where mixup methods have shown promise, such as speech and text classification. Our code is publicly available at: \url{https://github.com/wxitxy/synermix.git}.
\end{abstract}

\section{Introduction}

In the fields of deep learning and computer vision, the task of image classification stands as a fundamental and critical challenge. As datasets expand and model complexities increase, enhancing the generalization capabilities of models has become a focal point of research. Data augmentation is an effective technique that introduces variations into the training data to enrich its diversity, thereby improving the model's adaptability to novel data. Recently, data augmentation techniques such as MixUp \cite{bib1}, and its variants including Manifold MixUp \cite{bib2}, CutMix \cite{bib3}, and SaliencyMix \cite{bib4}, have been introduced as innovative strategies to improve the performance of image classification models. In practice, these augmentation methods are integrated into the training process governed by stochastic gradient descent (SGD). They work by pairing images or hidden representations from a shuffled version of a mini-batch with those from the original, unshuffled mini-batch. Each pair of corresponding images or hidden representations is then blended according to a mixing ratio to create synthesized images or representations.

\begin{figure}[!b]
\begin{adjustwidth}{-2.5cm}{-2.5cm}
\centering
\includegraphics[scale=.6]{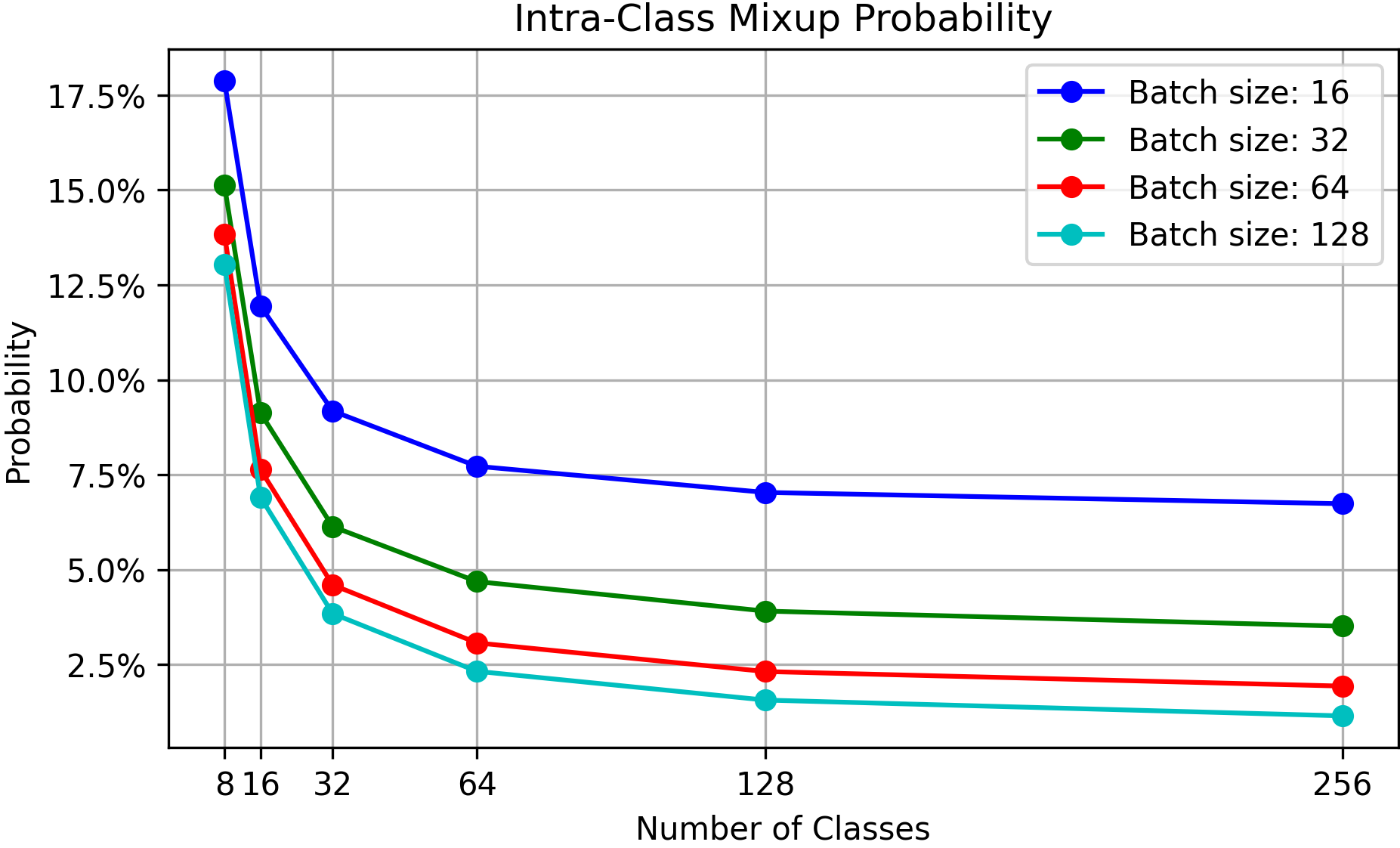}
\caption{\label{fig:fig1}Probability of occurrence for intra-class mixup implemented in existing mixup methods such as MixUp and Manifold MixUp.}
\end{adjustwidth}
\end{figure}

\begin{figure*}[!t]
\begin{adjustwidth}{-2.5cm}{-2.5cm}
	\centering
	\includegraphics[scale=.45]{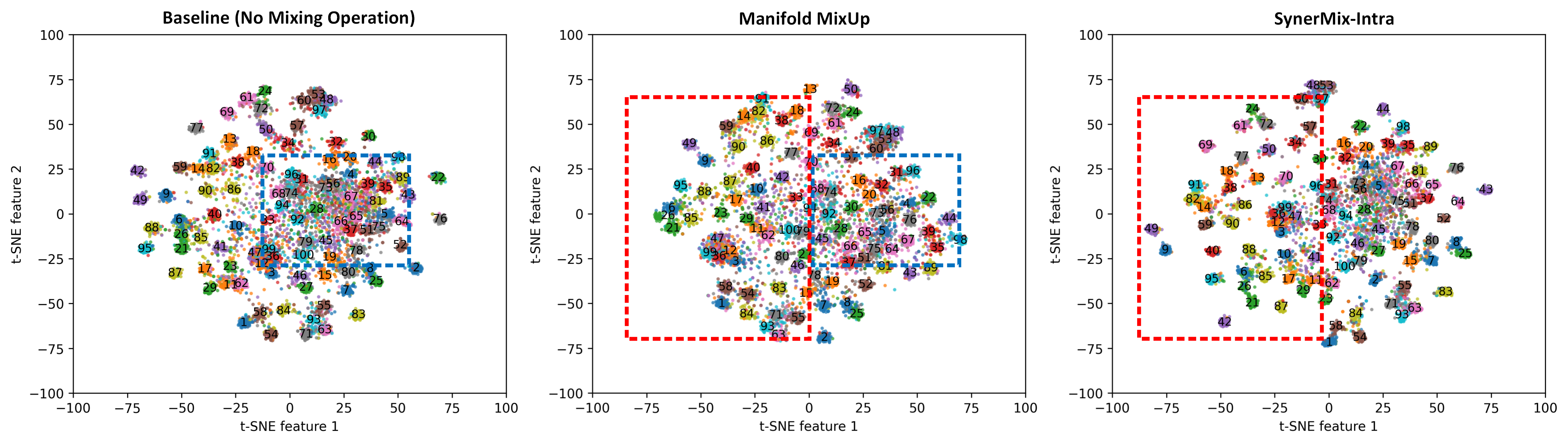}
	\caption{\label{fig:fig2}t-SNE visualization of feature representations for testing images from the \emph{CIFAR-100} dataset: comparing three methods – the baseline (no mixing operation), Manifold MixUp, and \emph{SynerMix-Intra}.}
\end{adjustwidth}
\end{figure*}

Despite the significant success achieved by existing mixup techniques, a closer examination reveals two critical limitations. The first concerns the mixing strategies—namely, intra-class and inter-class mixup. Existing mixup methods such as MixUp, Manifold MixUp, CutMix, and SaliencyMix often neglect intra-class mixup, which involves blending images or hidden representations from the same class. This oversight leads to an underutilization of the relationships among samples within the same class, potentially constraining image classification performance. The root of this issue lies in the shuffle operation; it results in a disproportionately high number of mixed pairs generated from different classes compared to those from the same class. Consequently, there is a predominant emphasis on inter-class mixup, which mixing images or hidden representations from different classes. As illustrated in Figure~\ref{fig:fig1}, with 16 classes and a mini-batch size of 32, the average probability of occurrence for intra-class mixup during a shuffle operation is only approximately 9\%. If the number of classes and the mini-batch size both increase to 128, this probability decreases to below 2\%. In this article, we refer to these techniques that primarily focus on inter-class mixup as ``inter-class mixup methods."

Additionally, the second limitation arises with respect to the mixing outcomes. Existing mixup methods, such as MixUp and Manifold MixUp, fall short in enhancing intra-class cohesion of image feature representations through mixing operations. In this article, the feature representation of an image refers to the input vector that is fed into the neural network's final classification layer—a fully connected layer responsible for predicting scores for each class. To be specific, the incidental intra-class mixup in current mixup methods indeed has the potential to facilitate the model's learning of common features among images from the same class, potentially leading to more cohesive clustering of feature representations. Nevertheless, the goal of forming compact class clusters is not explicit, and its effectiveness is significantly diminished due to the rarity of intra-class mixup. On the other hand, the predominant inter-class mixup they implement does enhance the distinction among different classes by promoting smoother predictions between classes but fails to offer any mechanism for reducing intra-class variability. As illustrated in Figure~\ref{fig:fig2}, especially within the region enclosed by the dashed blue box, the distinction among classes achieved by Manifold MixUp is noticeably improved compared to the baseline devoid of any mixing operations. However, the feature distribution of each class does not show a marked increase in clustering compared to the baseline. This is particularly obvious in comparison to the enhanced cohesion achieved by our mixup method named \emph{SynerMix-Intra} proposed in this article, as is evident within the red dashed box. In classification tasks, both intra-class cohesion and inter-class separability are known to be vital and equally significant. Consequently, the lack of a dedicated mechanism to improve intra-class cohesion in existing mixup methods may limit their performance in classification tasks.

To overcome these limitations, we propose the novel mixup method \emph{SynerMix-Intra} that significantly enhances intra-class cohesion through targeted intra-class mixup. Furthermore, we introduce a synergistic mixup solution named \emph{SynerMix} that effectively combines the benefits of both \emph{SynerMix-Intra} and existing mixup methods, addressing the two critical limitations identified in current mixup techniques.

\emph{SynerMix-Intra} focuses on intra-class mixup, an aspect that existing mixup techniques overlook, with the aim of enhancing intra-class cohesion—a benefit not provided by current mixup techniques. Specifically, during training, it generates a single synthesized feature representation for each class in every mini-batch. This synthesis is achieved through random linear interpolation among the feature representations of unaugmented original images within the same class. After the synthesized feature representations for all classes in the mini-batch are generated, they are fed into the final classification layer to produce predictions. An intra-class mixup loss is then calculated by averaging the classification losses of these synthesized representations. This loss is designed to promote the model to cluster image feature representations from the same class more tightly, resulting in enhanced intra-class cohesion at the end of training.

Building upon \emph{SynerMix-Intra} and existing mixup approaches, we have developed the synergistic mixup solution \emph{SynerMix}. \emph{SynerMix} leverages an existing mixup method—such as MixUp or Manifold MixUp—as the inter-class mixup component, while \emph{SynerMix-Intra} serves as the intra-class counterpart. The intra- and inter-class mixup components work together in a complementary manner, with the final loss calculated as a weighted sum of their losses, modulated by a balancing hyperparameter \(\beta\). In doing so, \emph{SynerMix} not only uniquely combines both inter- and intra-class mixup strategies in a balanced manner, but also simultaneously improves both intra-class cohesion and inter-class separability. It is capable of adapting to the distinct commonalities within classes and differences among classes that are inherent in various image classification tasks, thereby enhancing the model's generalization capabilities and robustness. Significantly, \emph{SynerMix} allows for flexible substitution of the inter-class mixup component; numerous existing mixup methods, including image-level mixup techniques like MixUp and feature-level mixup techniques like Manifold MixUp, can be implemented as the inter-class mixup component.

It is worth noting that \emph{SynerMix} does not deliberately exclude the occasional intra-class mixup that occurs in the existing mixup method, but rather integrates the entire method without any modifications. The reason is that the occasional intra-class mixup does not conflict with  \emph{SynerMix-Intra} and provides modest additional benefits, such as increased diversity in training data and enhanced feature robustness.

\emph{SynerMix} has been extensively validated through experiments on six public image datasets, including \emph{CIFAR-100} \cite{bib5}, \emph{Food-101} \cite{bib6}, \emph{mini-Imagenet} \cite{bib7}, \emph{OxfordIIIPet} \cite{bib8}, and \emph{Caltech-256} \cite{bib9}. MixUp and Manifold MixUp are selected as two implementations of the inter-class mixup component as they are seminal works in image-level and feature-level inter-class mixup, respectively. Experimental results demonstrate that the exclusive application of \emph{SynerMix-Intra} yields an accuracy gain ranging from 0.22\% to 3.09\% (averaging at 1.25\%) in 15 out of 18 experimental scenarios over the baseline devoid of any mixup operations. Furthermore, \emph{SynerMix} surpasses the best-performing method of either MixUp or \emph{SynerMix-Intra} by an accuracy margin of 0.1\% to 3.43\%, with an average improvement of 1.16\%. It also exceeds the top performer of either Manifold MixUp or \emph{SynerMix-Intra} by 0.12\% to 5.16\%, with an average gain of 1.11\%. These experimental results validate the effectiveness of \emph{SynerMix} in concurrently improving intra-class cohesion and inter-class separability. Additionally, given that \emph{SynerMix} is model-agnostic, it holds a substantial opportunity for application in other domains like speech and text classification \cite{bib1, bib10} where MixUp or its variants have proven useful.

The key contributions of our work include:
\begin{enumerate}
\item We have identified and underscored two critical limitations inherent in established mixup approaches such as MixUp and Manifold MixUp: a) neglect of intra-class mixup, leading to the underutilization of relationships among samples within the same class, and b) inadequacy in improving intra-class cohesion through their mixing operations, limiting image classification performance.

\item We have proposed a significant enhancement to existing mixup techniques with our mixup method \emph{SynerMix-Intra}. \emph{SynerMix-Intra} effectively fortifies intra-class cohesion through intra-class mixing operations. This enhancement addresses the deficiencies found in numerous mixup approaches, providing an important complement to these methods. To the best of our knowledge, \emph{SynerMix-Intra} represents the first attempt into leveraging mixing operations to specifically improve intra-class cohesion.

\item We have developed the synergistic mixup solution \emph{SynerMix} specifically designed to augment both intra-class cohesion and inter-class separability through mixing operations. This approach effectively overcomes the two inherent limitations of existing mixup methods and represents a pioneering effort to harness mixing operations for these dual objectives.

\item We have found a new method for incorporating beneficial stochasticity into gradients. This method stands apart from existing methods such as random mini-batches, dropout, weight regularization, and data augmentation. Our findings suggest that the randomness of synthesized feature representations within local feature spaces contributes to gradient stochasticity and can enhance the accuracy of image classification. This discovery opens up a new avenue for research: exploring the impact of synthesized feature representation randomness on gradient stochasticity as a means for improving image classification performance.
\end{enumerate}

The structure of this paper is as follows: Section 2 reviews related work. Section 3 details the synergistic mixup solution \emph{SynerMix}. Section 4 presents the experimental results. Finally, Section 5 concludes this article.

\section{Related Work}
Our work contributes to the field of image data augmentation techniques, a domain that is pivotal for the advancement of image classification performance. Traditional data augmentation methods aim to artificially expand the training dataset by applying a series of transformations that produce varied but plausible versions of training images. Common transformations \cite{bib11} include geometric modifications such as rotation, translation, scaling, and flipping, which help models to become invariant to changes in the position and orientation of objects within images.

As the field of data augmentation has evolved, researchers have explored beyond traditional transformation-based techniques to more sophisticated approaches. Generative models, such as Generative Adversarial Networks (GANs) \cite{bib12} and Variational Autoencoders (VAEs)  \cite{bib13}, have been at the forefront of this innovation, synthesizing entirely new images that closely adhere to the distribution of the training data. Neural style transfer \cite{bib14} and feature space augmentation \cite{bib15} represent further advancements, introducing stylistic variations and promoting model robustness against more abstract input variations. While generative models and neural style transfer expand the horizons of data augmentation, they often come with increased computational demands and complexity. In contrast, feature space augmentation, particularly through simple operations such as adding noise, applying feature dropout, and performing interpolation, remains computationally efficient and less complex.

Besides these advanced techniques, MixUp has gained significant attention due to its simplicity and effectiveness. It involves generating synthesized examples by linearly blending pairs of training samples and their corresponding labels, encouraging the model to predict more smoothly between classes. Several MixUp variants have emerged and been applied to various domains, such as image classification, regression (e.g., C-Mixup \cite{bib16}), graph deep learning model training (e.g., G-Mixup \cite{bib17}), speech recognition (e.g., MixUp), text classification (e.g., wordMixup \cite{bib18}, senMixup \cite{bib18}, Mixup-Transformer \cite{bib10}), and action recognition (e.g., \cite{bib19}).

In the following subsections, we provide a literature review of mixup techniques related to image classification, categorizing them from the perspective of mixing strategies, specifically inter-class and intra-class mixup.

\subsection{Inter-Class Mixup Technique}
Existing inter-class mixup approaches involve mixing images or hidden representations. We categorize these into image-level inter-class mixup and feature-level inter-class mixup, as detailed below.

\subsubsection{Image-Level Inter-Class Mixup}
In addition to the seminal MixUp technique, several extensions are also dedicated to performing mixing operations at the image level. CutMix \cite{bib3} avoids the issue of unnatural blending often seen in MixUp by dropping a region from one image and filling it with a patch from another, creating a mixed image. To resolve the ``strong-edge" artifacts that can arise in the areas where the patch is applied, SmoothMix \cite{bib20} introduces the use of soft edges for a smoother transition between the combined image regions.

In order to address the issue of CutMix that random selection strategies of the patches may not necessarily represent sufficient information about the corresponding object, SaliencyMix \cite{bib4} carefully selects a representative image patch using a saliency map and mixes this indicative patch with the target image. PuzzleMix \cite{bib21} also focuses on ensuring that the mixed image contains salient information from both source images. It seeks to find the optimal mask that determines the proportion of each input to reveal or conceal within a given region, and the optimal transport that maximizes the exposed saliency under the mask.

To solve the manifold intrusion issue associated with MixUp, AdaMixUp \cite{bib22} introduces two neural networks to generate the mixing coefficient and to assess whether the newly mixed data causes manifold intrusion, addressing the issue where Mixup-generated samples may closely resemble existing real samples but with differing label semantics.

\subsubsection{Feature-Level Inter-Class Mixup}
Beyond the image level, some researchers have explored the application of MixUp to the feature space. A notable work is Manifold MixUp \cite{bib2}, proposed by Verma et al., which extends the mixup concept to the latent spaces of neural networks. This method blends hidden representations of two images at various hidden layers, promoting smoother decision boundaries at different levels of abstraction. FC-mixup \cite{bib23} adopts a CutMix-inspired approach to mix the feature volumes of two images, where some feature maps or individual pixels within the feature volume of one image are removed and replaced with corresponding feature maps or pixels from another image's feature volume. 

Feature-level mixup has been found capable of solving more complicated problems such as domain generalization, long-tailed classification and object detection. For domain generalization, MixStyle \cite{bib24} blends the feature statistics of two images to create new domains. This method is inspired by insights from style transfer research, which suggests that feature statistics capture style- or domain-related information. Also, FIXED \cite{bib25} enhances MixUp for domain generalization by integrating domain-invariant learning and introducing a larger margin to reduce synthetic noise, addressing MixUp's limitations in discerning domain-class information and improving classifier robustness. For long-tailed classification, MixUp and Mainfold MixUp have shown performance improvements on long-tailed datasets. To further boost their performance, AM-mixup \cite{bib26} strategically controls the mixup ratio to generate sythesis features that asymptotically approach the midpoint between classes, thereby balancing class margins and mitigating inter-class and intra-class collapse effects. For object detection, Chen et al. \cite{bib27} proposed a domain generalization approach for underwater object detection, employing style transfer to diversify training domains, feature-level mixup for interpolating domain features, and contrastive learning for invariant feature extraction.

\subsection{Intra-class Mixup Technique}
The majority of methods, including MixUp, CutMix, SmoothMix, SaliencyMix, PuzzleMix, AdaMixUp, Manifold MixUp, FC-mixup, MixStyle, FIXED, and AM-mixup, involve intra-class mixup. However, as demonstrated in Figure~\ref{fig:fig1}, the likelihood of occurrence for it diminishes with larger mini-batch sizes and more classes, making it less significant compared to the inter-class mixup they implement.

Beyond these approaches, a few studies have recognized the importance of intra-class mixup. Intra-Class CutMix, as discussed in \cite{bib28}, addresses class imbalance by blending images from minority classes, which enhances the performance of neural networks on unbalanced datasets. In \cite{bib29}, intra-class mixing of images is leveraged to advance out-of-distribution (OoD) detection by minimizing the angular dispersion in latent space representations, thereby sharpening the demarcation between in-distribution and OoD data. AugMix \cite{bib30} merges randomly produced augmentations of a single image and utilizes a Jensen-Shannon divergence-based loss to promote consistency. Intra-class Part Swapping (InPS) \cite{bib31} generates new training data by performing attention-guided content swapping on input pairs from the same class. It avoids introducing noisy labels and ensures a reasonable holistic structure of objects in generated images.\\

After reviewing the works mentioned above, it becomes evident that they inherently possess the following two significant limitations. The first is the neglect of intra-class mixup in contrast to inter-class mixup. The works presented in Section 2.1, as well as those applied to other domains beyond image classification (e.g., \cite{bib10, bib18, bib19}), emphasize mixing between different classes while neglecting mixing within the same class. Although a few studies have specifically focused on intra-class mixup, they do not fully recognize its critical role in reducing intra-class variability. Furthermore, they do not leverage the combination of intra-class and inter-class mixup to maximize the model's classification performance. The second limitation is the lack of augmenting the cohesion of image feature representations within the same class through their mixing operations. While these mixup methods succeed in improving inter-class separability via inter-class mixup, they do not address the enhancement of intra-class cohesion—a factor that is also crucial for image classification. To bridge these gaps, this article proposes the novel mixup method \emph{SynerMix-Intra} that strengthens the cohesion of intra-class feature representations, as well as the synergistic mixup solution \emph{SynerMix} that improves both intra-class cohesion and inter-class separability.

\section{Proposed Method}
\subsection{Methodology}
\begin{figure*}[t]
\begin{adjustwidth}{-2.5cm}{-2.5cm}
    \centering
    \includegraphics[width=0.85\paperwidth]{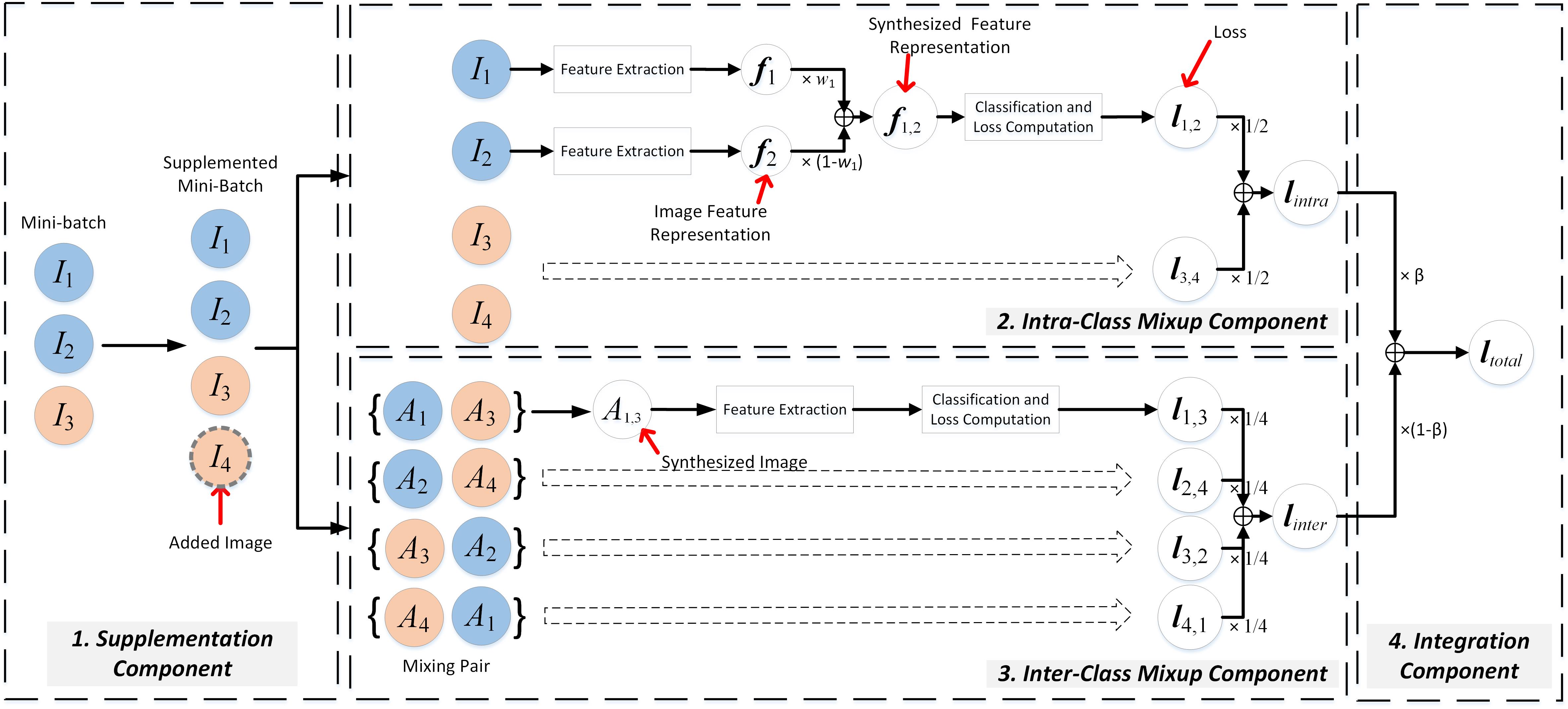}
    \caption{\label{fig:fig3}Pipeline of loss calculation for \emph{SynerMix}. It consists of four components including the supplementation component, intra-class mixup component, inter-class mixup component and integration component. $I_1$ through $I_4$ denote the four unaugmented original images, and $A_1$ through $A_4$ represent the augmented images obtained from $I_1$ through $I_4$ by applying a series of augmentation operations such as cropping, flipping, and cutout. Blue and orange colors are used to differentiate the two classes. Notably, many established mixup methods, such as MixUp and Manifold MixUp, can be used to implement the inter-class mixup component. Here, we only present the loss calculation process of MixUp for clarity.}
\end{adjustwidth}
\end{figure*}

In the context of using data augmentation techniques with SGD, the loss calculation process of \emph{SynerMix} for a mini-batch is depicted in Figure~\ref{fig:fig3}. This process comprises four components:
\begin{itemize}
\item Supplementation Component: Ensure that each class within the mini-batch contains at least two images.

\item Intra-Class Mixup Component: Generate synthesized feature representations by blending feature representations of unaugmented original images from each class and calculate the average classification loss \(\mathscr{L}_{\text{intra}}\).

\item Inter-Class Mixup Component: Blend augmented images or their hidden representations between different classes and compute the average classification loss~\(\mathscr{L}_{\text{inter}}\).

\item Integration Component: Combine the losses from the intra-class and inter-class components, modulated by a balancing hyperparameter \(\beta\).
\end{itemize}
In the following, we describe each component in detail.

\subsubsection{Supplementation Component}
\emph{SynerMix-Intra}, as the unique implementation for the intra-class mixup component, requires interpolating image feature representations from the same class within each mini-batch. This process requires that there be at least two images for each class contained in the mini-batch to facilitate the operation. SGD involves randomly sampling each mini-batch from the training dataset, which does not ensure that every class within the mini-batch will have a minimum of two images. Therefore, it becomes crucial to augment each mini-batch to guarantee that for every class appearing in the mini-batch, there are at least two images available.

To address this, we devise a specific supplementation strategy for mini-batches. Specifically, for any class represented by only one sample in a mini-batch, an additional image of the same class from the training set (excluding those already in the mini-batch) is randomly selected and added to the mini-batch. This supplementation process increases the size $N$ of the mini-batch, particularly when the mini-batch size is significantly smaller than the number of classes. In extreme cases, where each class is represented by only one image in the mini-batch, this process will double the size of the mini-batch.

\subsubsection{Intra-class Mixup Component}
\emph{SynerMix-Intra} is the exclusive implementation for the intra-class mixup component. Its details are as follows. For a given supplemented mini-batch sequence {($I_1$, $y_1$), \ldots, ($I_{\hat{N}}$, $y_{\hat{N}}$)} where $y_i$ represents the class label of the unaugmented original image $I_i$ and $\hat{N}$ the supplemented mini-batch size, we input \(I_1, ..., I_{\hat{N}}\) into the model and obtain the vectors \(\bm{f}_1, ..., \bm{f}_{\hat{N}}\) fed into the final classification layer. These vectors serve as the feature representations of these images. The single synthesized feature representation $\bm{f}_{\text{syn}}^c$ for the class labeled $c$ is defined as:

\begin{equation}
\bm{f}_{\text{syn}}^c = \sum_{y_j=c} \omega_j \bm{f}_{j} \quad \text{s.t. } \sum_{y_j=c} \omega_j = 1; \omega_j \geq 0
\end{equation}
where $\omega_j$ represents the interpolation weight for the feature $\bm{f}_j$. The weight $\omega_j$ is derived from the random number $r_j$ sampled from a uniform distribution $U(0, 1)$, and is normalized with respect to the sum of random numbers $r_i$ associated with all images belonging to class $c$ within the mini-batch. Mathematically, $\omega_j$ is expressed as $\omega_j = r_j / \sum_i r_i$.

Assuming that the mini-batch contains a total of $C$ classes, the intra-class mixup loss for the mini-batch is given by:

\begin{equation}
\mathscr{L}_{\text{intra}} = \frac{1}{C} \sum_{c=1}^{C} \ell(\bm{f}_{\text{syn}}^c, \boldsymbol{t}_c)
\end{equation}
where $l(\cdot, \cdot)$ denotes the loss function and $\boldsymbol{t}_c$ is the one-hot encoded label vector for class $c$.

\subsubsection{Inter-class Mixup Component}
Many existing mixup methods, such as MixUp, ManifoldMixup, CutMix, SmoothMix, SaliencyMix, PuzzleMix, AdaMixUp, FC-mixup, MixStyle, FIXED, and AM-mixup, can be directly used to implement the inter-class mixup component without any modifications. Here, we briefly revisit MixUp and Manifold MixUp since they are seminal works in image-level and feature-level inter-class mixup.

For a given supplemented mini-batch sequence \{($A_1$, $\bm{y}_1$), \ldots, ($A_{\hat{N}}$, $\bm{y}_{\hat{N}}$)\} where $A_i$ represents the augmented image derived from the unaugmented original image $I_i$ and $\bm{y}_i$ is the one-hot encoded label vector of the class label $y_i$, we obtain a new sequence \{($A_1'$, $\bm{y}_1'$), \ldots, ($A_{\hat{N}}'$, $\bm{y}_{\hat{N}}'$)\} after shuffling. We then pair the two sequences to form pairs \{[($A_1$, $\bm{y}_1$), ($A_1'$, $\bm{y}_1'$)], \ldots, [($A_{\hat{N}}$, $\bm{y}_{\hat{N}}$), ($A_{\hat{N}}'$, $\bm{y}_{\hat{N}}'$)]\}.

With the function $\text{Mix}_{\lambda}(a, b) = \lambda \cdot a + (1 - \lambda) \cdot b$, which mixes $a$ and $b$ according to $\lambda$ drawn from a Beta distribution $\text{Beta}(\alpha, \alpha)$ (when $\alpha = 1.0$, this is equivalent to sampling $\lambda$ from a uniform distribution $U(0, 1)$), the loss $\mathscr{L}_{\text{inter}}$ when applying MixUp is given by:

\begin{equation}
\mathscr{L}_{\text{inter}} = \frac{1}{\hat{N}} \sum_{i=1}^{\hat{N}} \ell(\text{Mix}_{\lambda_i}(A_i, A_i'), \text{Mix}_{\lambda_i}(\bm{y}_i, \bm{y}_i'))
\end{equation}
where $\ell(\cdot, \cdot)$ is the loss function, and $\lambda_i$ is drawn from a Beta distribution $Beta(\alpha, \alpha)$ for the $i$-th image pair.

The loss $\mathscr{L}_{\text{inter}}$ when applying Manifold MixUp is given by:
\begin{equation}
\mathscr{L}_{\text{inter}} = \frac{1}{\hat{N}} \sum_{i=1}^{\hat{N}} \ell(\text{Mix}_{\lambda_i}(g_{k_i}(A_i), g_{k_i}(A_i')), \text{Mix}_{\lambda_i}(\bm{y}_i, \bm{y}_i'))
\end{equation}
where $g_{k_i}(A_i)$ and $g_{k_i}(A_i')$ are the outputs of the neural network at the layer $k_i$ for $A_i$ and $A_i'$, respectively. The layer $k_i$ is randomly selected from a set of eligible layers $\mathcal{S}$ for the $i$-th image pair. When $\mathcal{S}$ contains only the input layer, then Manifold MixUp reduces to MixUp. 

\subsubsection{Integration Component}
The loss $\mathscr{L}_{\text{total}}$ of each mini-batch is composed of two parts: $\mathscr{L}_{\text{intra}}$ and $\mathscr{L}_{\text{inter}}$. The combined loss is calculated as follows:

\begin{equation}
\mathscr{L}_{\text{total}} = \beta \cdot \mathscr{L}_{\text{intra}} + (1 - \beta) \cdot \mathscr{L}_{\text{inter}}
\end{equation}
where $\beta$ is a hyperparameter between 0 and 1 that balances the contributions of the two types of losses. 

\subsection{Method Explanation}
This subsection thoroughly explains the rationale behind \emph{SynerMix} by addressing the following three pivotal questions. Additionally, we detail the development of the loss calculation formula, tracing its evolution from the initial formulation in Eq. (6) to its finalized version in Eq. (5). To provide a clear response to the first question, we assume that the feature representations of augmented images, rather than those of unaugmented original images, are employed to generate synthetic feature representations. In relation to the second question, we will explore the potential side effect of this assumption and examine why utilizing the feature representations of unaugmented original images to produce synthetic representations can alleviate this side effect.

\begin{figure}[!t]
\begin{adjustwidth}{-2.5cm}{-2.5cm}
\centering
\includegraphics[scale=.85]{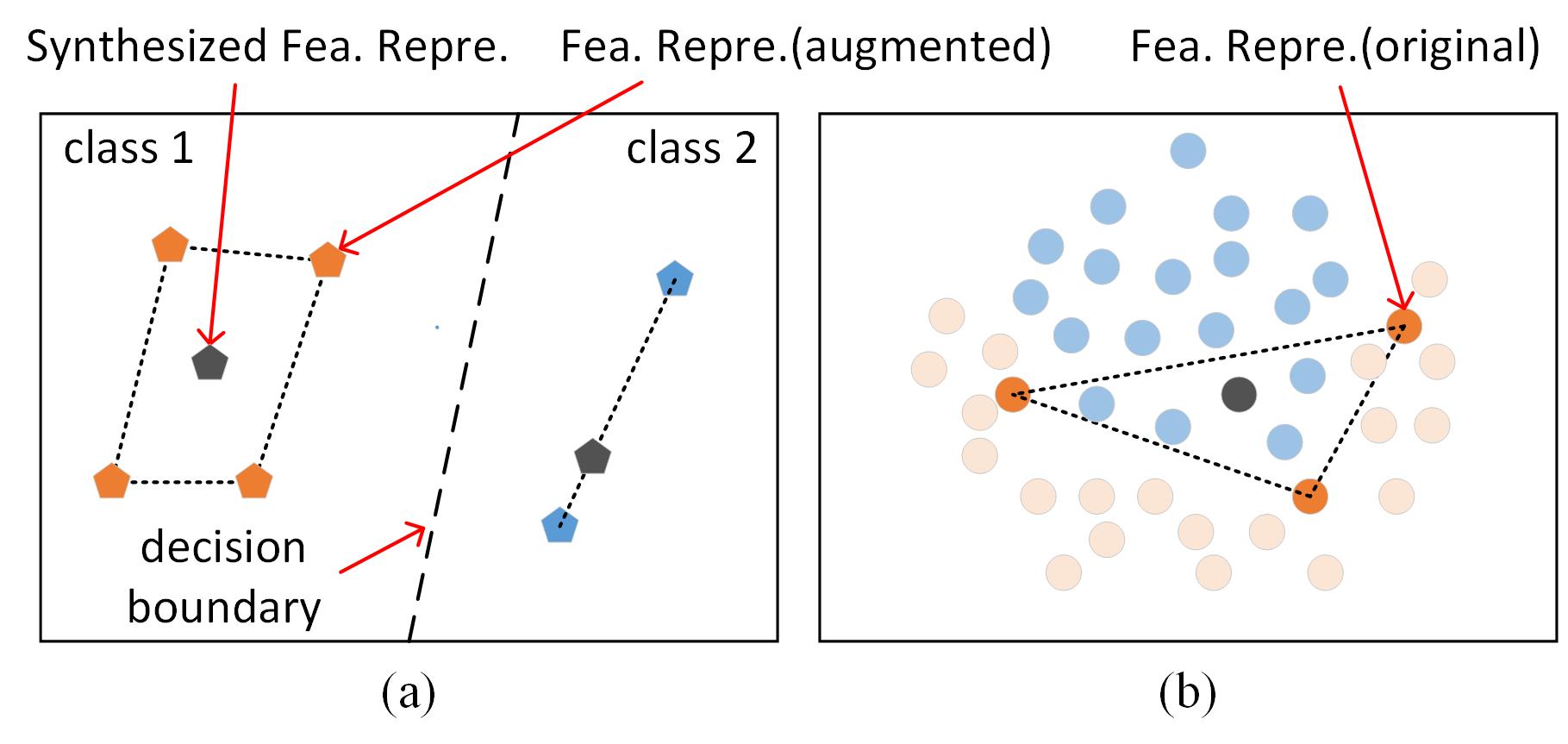}
\caption{\label{fig:fig4}Toy examples of feature representation synthesis: (a) Illustration of enhancing the cohesion of image feature representations within the same class; (b) Illustration of using feature representations from unaugmented original images for synthesis.}
\end{adjustwidth}
\end{figure}

\textbf{(1) Why can \emph{SynerMix-Intra} improve intra-class cohesion?} As shown in Figure~\ref{fig:fig4}(a), each randomly synthesized feature representation can be viewed as a point within the convex hull or on a line segment formed by the feature representations of augmented images that generate it. Moreover, the last fully-connected layer, which receives image feature representations as input, acts as a linear classifier. In this case, given the requirement that any point within the geometric space formed by feature representations of augmented images from the same class must be correctly classified, these feature representations are encouraged to cluster more tightly together, thereby expediting the fulfillment of this requirement. Crucially, the presence of this encouragement is consistent throughout the training process. As a result, compared to the scenario where this encouragement is absent, there is not only an enhancement in intra-class cohesion upon completion of training, but also a more rapid aggregation of same-class image feature representations during the training process, leading to consistently higher classification accuracy, as demonstrated in Section 4.2.1. In the implementation of intra-class mixup, we ensure that each weight \(w_j \geq 0\), which guarantees that the synthesized feature representations lie within the formed geometric space. By generating \(w_j\) via random numbers sampled from a uniform distribution $U(0, 1)$, we ensure the complete randomness of synthesized representations within the formed geometric space.

In the above scenario, the loss calculation formula can be derived from the empirical risk minimization (ERM) framework:

\begin{equation}
\mathscr{L}_{\text{total}} = \mathscr{L}_{\text{augment}}
\end{equation}
to:
\begin{equation}
\mathscr{L}_{\text{total}} = \beta \cdot \mathscr{L}_{\text{intra(aug)}} + (1 - \beta) \cdot \mathscr{L}_{\text{augment}}
\end{equation}
where $\mathscr{L}_{\text{intra(aug)}}$ is obtained by mixing feature representations of augmented images instead of unaugmented original images, and $\mathscr{L}_{\text{augment}}$ is calculated as $\frac{1}{\hat{N}} \sum_{i=1}^{\hat{N}} \ell(A_i, \bm{y}_i)$, where no mixup operation is performed.

\textbf{(2) Why is it beneficial to use feature representations from unaugmented original images in order to generate synthesized ones?} The reasonableness of expecting synthesized feature representations to be correctly classified hinges on the assumption that they accurately reflect the true distribution. However, this assumption does not always hold. Figure~\ref{fig:fig4}(b) clearly demonstrates that even when a synthesized representation is generated from authentic feature representations, it does not belong to the true distribution. Furthermore, the negative impact will be amplified when it overlaps with the feature distribution of another class.

Employing augmented images for synthesized representation generation will intensify the issue. Augmentation techniques have the potential to alter key semantic details—for instance, through improper region cutouts—resulting in feature representations that diverge from the true distribution. Synthesized representations based on such altered feature representations are more prone to straying from the true distribution. Consequently, we opt to synthesize feature representations using those from unaugmented original images instead of augmented images.  In this context, Eq. (7) evolves into the following form:
\begin{equation}
\mathscr{L}_{\text{total}} = \beta \cdot \mathscr{L}_{\text{intra}} + (1 - \beta) \cdot \mathscr{L}_{\text{augment}}
\end{equation}
where $\mathscr{L}_{\text{intra}}$ is the intra-class mixup loss defined in Eq. (2), and $\mathscr{L}_{\text{augment}}$ is the loss calculated for augmented images as defined in Eq. (7).

\textbf{(3) What are the benefits of combining intra-class mixup with inter-class mixup?} In \emph{SynerMix}, the intra-class mixup component works synergistically with the inter-class mixup component. The former ensures that image feature representations of the same class are more tightly clustered. Meanwhile, the latter increases inter-class separability, making different classes more distinguishable from one another. The seamless integration of these methods yields a comprehensive loss function that incorporates both intra- and inter-class mixup strategies in a harmonious balance. Consequently, this dual integration serves to concurrently improve intra-class cohesion and inter-class separability. Moreover, the introduction of the hyperparameter \(\beta\) allows for an adjustable balance, which is essential for training a model that is well-suited for the complexities of image classification. This adaptability enables the model to account for the unique commonalities within classes and distinctions among classes inherent in different image classification tasks. Building on these benefits, Eq. (8) is augmented by substituting its \(\mathscr{L}_{\text{augment}}\) with \(\mathscr{L}_{\text{inter}}\), culminating in the final loss equation presented in Eq. (5).

\section{Experiments}

\begin{table}[htbp]
\small
\begin{adjustwidth}{-2.5cm}{-2.5cm}
\centering
\caption{Datasets and Training Parameters. rc: random cropping, hf: random horizontal flipping, co: cutout}
\label{tab:datasets}
\begin{tabular}{@{}l|p{2cm}|p{2.5cm}|p{2.6cm}|p{3.4cm}|p{3.4cm}@{}}
\toprule
Dataset & CIFAR-100 & Food-101 & mini-ImageNet & Caltech-256 & OxfordIIIPet \\ 
\midrule
Classes & 100 & 101 & 100 & 257 & 37 \\
Training Images/Class & 500 & 750 & 500 & 60 & 100 \\
Model Input Size & $32 \times 32$ & $224 \times 224$ & $84 \times 84$ & $224 \times 224$ & $224 \times 224$ \\
Initial Learning Rate & 0.1 & 0.1 & 0.1 & 0.01 & 0.01 \\
Momentum,Weight Decay & 0.9, 0.0005 & 0.9, 0.0005 & 0.9, 0.0005 & 0.9, 0.0005 & 0.9, 0.0005 \\
Step Size,Gamma & 10, 0.5 & 10, 0.5 & 5, 0.5 & 5, 0.5 & 5, 0.5 \\
Total Epochs & 120 & 120 & 60 & 60 & 60 \\
Data Augmentation & rc, hf, co & rc, hf & rc, hf, co & rc, hf & rc, hf \\
Pre-trained & No & No & No & Yes & Yes \\
Model & ResNet-18, ResNet-34 & ResNet-50, ResNet-101 & ResNet-18, ResNet-34 & tiny-SwinTransformer, MobileNetV3-Large, ResNet-50 &tiny-SwinTransformer,  MobileNetV3-Large, ResNet-34 \\

\bottomrule
\end{tabular}
\end{adjustwidth}
\end{table}

\subsection{Experimental Setup}
\subsubsection{Training Setup}
In our experiments, we have meticulously curated a diverse experimental setup to effectively assess the robustness and generalizability of \emph{SynerMix}. This diversity is reflected in the selection of datasets, model input sizes, model architectures, and the use of pre-trained models.

\begin{itemize}
\item Dataset Diversity: We have chosen a range of datasets, including \emph{CIFAR-100}, \emph{Food-101}, \emph{mini-ImageNet}, \emph{Caltech-256}, and \emph{OxfordIIIPet}, to cover a broad spectrum of image classification tasks. These datasets vary significantly in terms of the number of classes, from 37 in \emph{OxfordIIIPet} to 257 in \emph{Caltech-256}. This variety ensures that our findings are indicative of model performance across a range of scenarios from general object classification (\emph{CIFAR-100}, \emph{mini-ImageNet}, \emph{Caltech-256}) to specific domain fine-grained classification (\emph{Food-101}, \emph{OxfordIIIPet}).

\item Model Input Size Diversity: Our selection of datasets incorporates a variety of model input sizes, from \emph{CIFAR-100}'s $32 \times 32$ pixels to the $224 \times 224$ pixels of \emph{Food-101}, \emph{Caltech-256}, and \emph{OxfordIIIPet}, with \emph{mini-ImageNet} at an intermediate $84 \times 84$ pixels. This variation tests the performance of \emph{SynerMix} across different input sizes.

\item Model Architecture Diversity: We have selected a range of model architectures, including the less complex ResNet-18 \cite{bib32}, the moderately complex ResNet-34 and ResNet-50 \cite{bib32}, the deeper ResNet-101 \cite{bib32}, the efficient MobileNetV3-Large \cite{bib33}, and the compact but advanced tiny-SwinTransformer \cite{bib34}. This lineup allows us to evaluate \emph{SynerMix}'s adaptability across models with varying complexities and depths.

\item Utilization of Pre-trained Models: Our study investigates whether our method is effective when models are trained from scratch as well as when they are initialized with weights from pre-trained models. This is to validate that \emph{SynerMix} is robust and functions well under both conditions.
\end{itemize}

Table 1 shows the complete training setup. For \emph{CIFAR-100}, we trained ResNet-18 and ResNet-34 models from scratch for 120 epochs, using SGD with a learning rate of 0.1, momentum of 0.9, and weight decay of $5 \times 10^{-4}$. The learning rate was reduced by half every 10 epochs. Similarly, for \emph{Food-101}, \emph{ResNet-50} and \emph{ResNet-101} models were trained from scratch for 120 epochs with identical SGD parameters and learning rate adjustments.

For \emph{mini-ImageNet}, the training setup was the same as \emph{CIFAR-100}, but the learning rate was halved every 5 epochs, and the models were trained for 60 epochs. For \emph{Caltech-256} and \emph{OxfordIIIPet}, we fine-tuned ResNet-50 and tiny-SwinTransformer models that were pre-trained on the \emph{ImageNet-1K} \cite{bib35} dataset. The fine-tuning began with a learning rate of 0.01, which was halved every 5 epochs, and the models were trained for a total of 60 epochs.

Data augmentation for \emph{CIFAR-100} and \emph{mini-ImageNet} included random cropping, horizontal flipping, and cutout \cite{bib36}, while for \emph{Food-101}, \emph{Caltech-256}, and \emph{OxfordIIIPet}, only random cropping and horizontal flipping were used.

\subsubsection{Comparable Methods}
To conduct a comprehensive evaluation, we designed five distinct methods that selectively employ int\emph{ra}-class mixup (denoted as \emph{RA}), int\emph{er}-class mixup (denoted as \emph{ER}), both, or neither, each with a specific focus. In the following, ``\emph{w-}" and  ``\emph{wo-}" denotes ``with" and ``without", respectively:

\begin{itemize}
\item \emph{wo-RA\&ER} directly trains on augmented images without any mixup operations. It adheres to the loss function outlined in Eq. (6).

\item \emph{w-ER} only applies inter-class mixup. This method is implemented through MixUp or Manifold MixUp, denoted by the suffix ``\emph{(M)}" or ``\emph{(MM)}", respectively. Its associated loss functions are detailed in Eq. (3) or Eq. (4) accordingly.

\item \emph{w-RA} represents our proposed mixup method \textbf{\emph{SynerMix-Intra}}. It performs intra-class mixup on the feature representations of unaugmented original images, with the loss function given by Eq. (8).

\item \emph{w-RA(aug)} applies intra-class mixup to the feature representations of augmented images. The corresponding loss function is detailed in Eq. (7).

\item \emph{w-RA\&ER} indicates our synergistic mixup solution \textbf{\emph{SynerMix}}, which combines \emph{w-RA} and \emph{w-ER}. Its loss function is Eq. (5).
\end{itemize}

In our experiments, MixUp and Manifold MixUp are selected as two implementations of the inter-class mixup component as they are representative of image-level and feature-level inter-class mixup, respectively. For these methods, the optimal setting for $\alpha$ varies across datasets and mini-batch sizes and can be equal to, greater than, or less than 1. Given the substantial computational cost associated with fine-tuning $\alpha$ for each specific scenario, we opt to set $\alpha$ to 1 by default. This standardization consistently results in a marked increase in accuracy in our experiments.

In addition, Manifold MixUp needs to select a layer's output for mixing at each mini-batch from a predetermined set $\mathcal{S}$ of candidate outputs. For the ResNet architecture, $\mathcal{S}$ encompasses the input image, as well as the outputs from the first and second residual blocks. For MobileNetV3-Large, $\mathcal{S}$ includes the input image, the output after the first hardswish activation function, and the outputs from the first and second inverted residual modules. For the tiny-SwinTransformer model, $\mathcal{S}$ comprises the input image, the output after the first layer normalization operation, and the outputs from the second and fourth transformer blocks. This strategic selection of outputs for mixing in each architecture consistently leads a noticeable accuracy improvement over MixUp across various datasets and mini-batch sizes, which aligns with the findings presented in \cite{bib2}.

\subsubsection{Valiation Setup and Evaluation Metrics}
During the validation stage, we allocated 10\% of the training data from each dataset to form a validation set through stratified sampling, ensuring that the class distribution in the validation set mirrors that of the original training data. The remaining 90\% of the data was used for training purposes.  The validation set played a crucial role in fine-tuning the hyperparameter $\beta$. For \emph{w-RA}, $\beta$ options were [0.2, 0.4, 0.5, 0.7], and for \emph{w-RA\&ER(M)} and \emph{w-RA\&ER(MM)}, choices were [0.05, 0.1, 0.2, 0.4]. In the testing stage, the entire training set is used to train models using the best $\beta$ from the validation stage. Given that each dataset's test set is balanced, we used accuracy as the sole metric for performance evaluation. To mitigate the effects of variability in model accuracy over epochs, we averaged the last 10 epochs for \emph{CIFAR-100} and \emph{Food-101} over 120 epochs, and the last 5 epochs for other datasets over 60 epochs. The effectiveness of \emph{SynerMix} was evaluated in achieving higher classification accuracy compared to when \emph{w-RA} and \emph{w-ER} are used independently. 

\begin{figure*}[!t]
\begin{adjustwidth}{-2.5cm}{-2.5cm}
	\centering
	\begin{minipage}[t]{0.42\linewidth}
		\includegraphics[width=\linewidth]{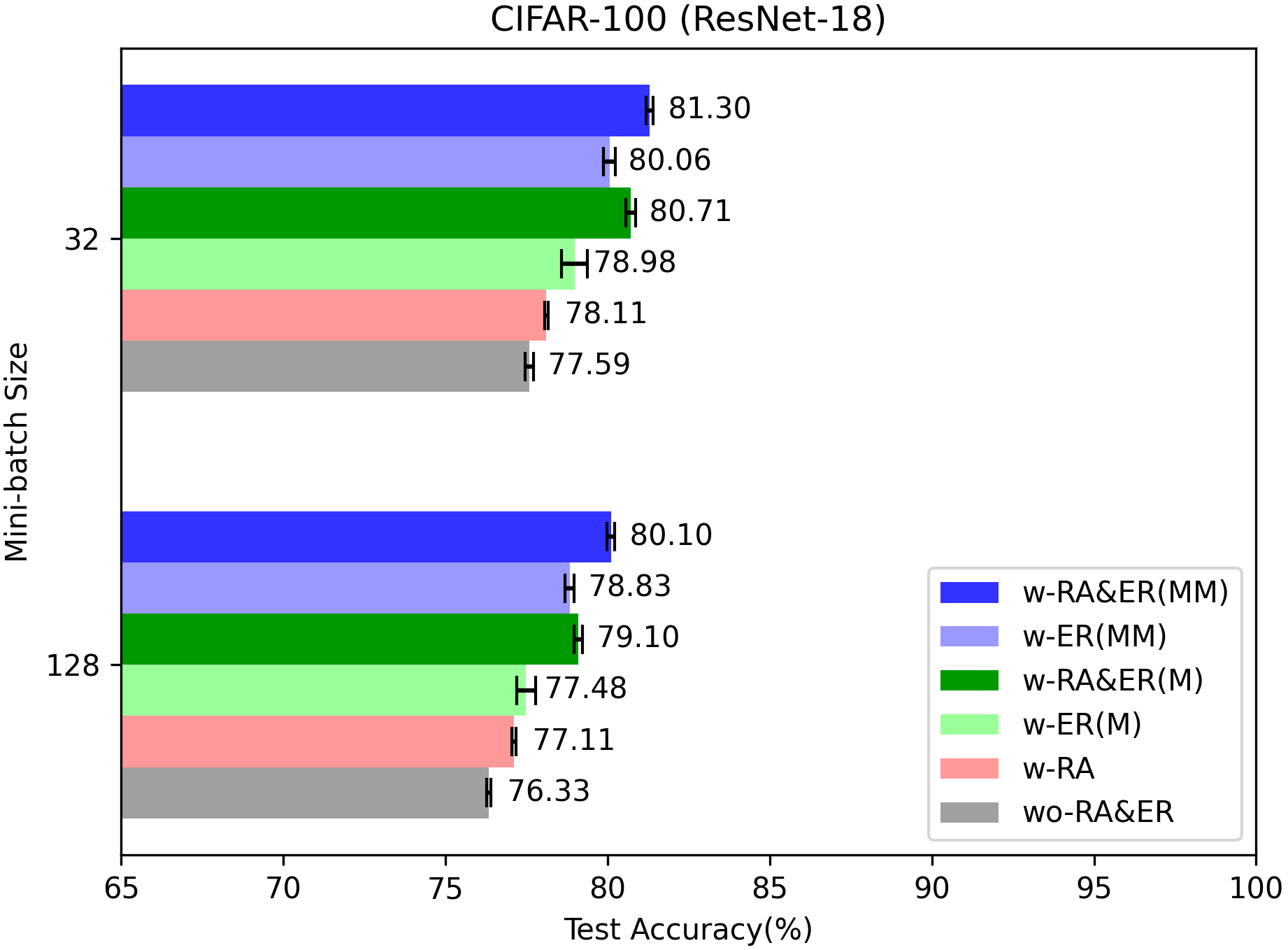}
	\end{minipage}%
 	\begin{minipage}[t]{0.42\linewidth}
		\includegraphics[width=\linewidth]{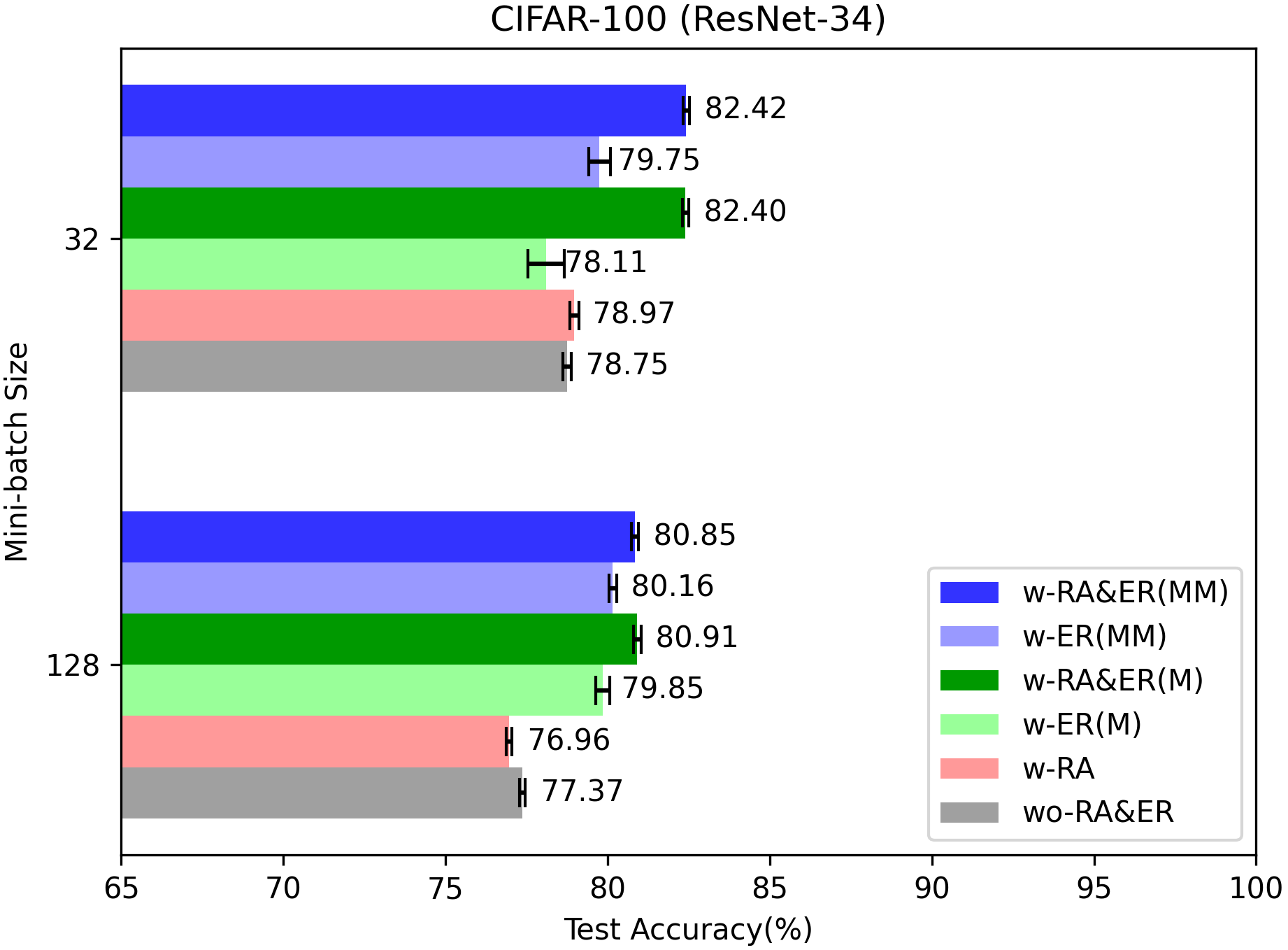}
	\end{minipage}%
 
	\begin{minipage}[t]{0.42\linewidth}
		\includegraphics[width=\linewidth]{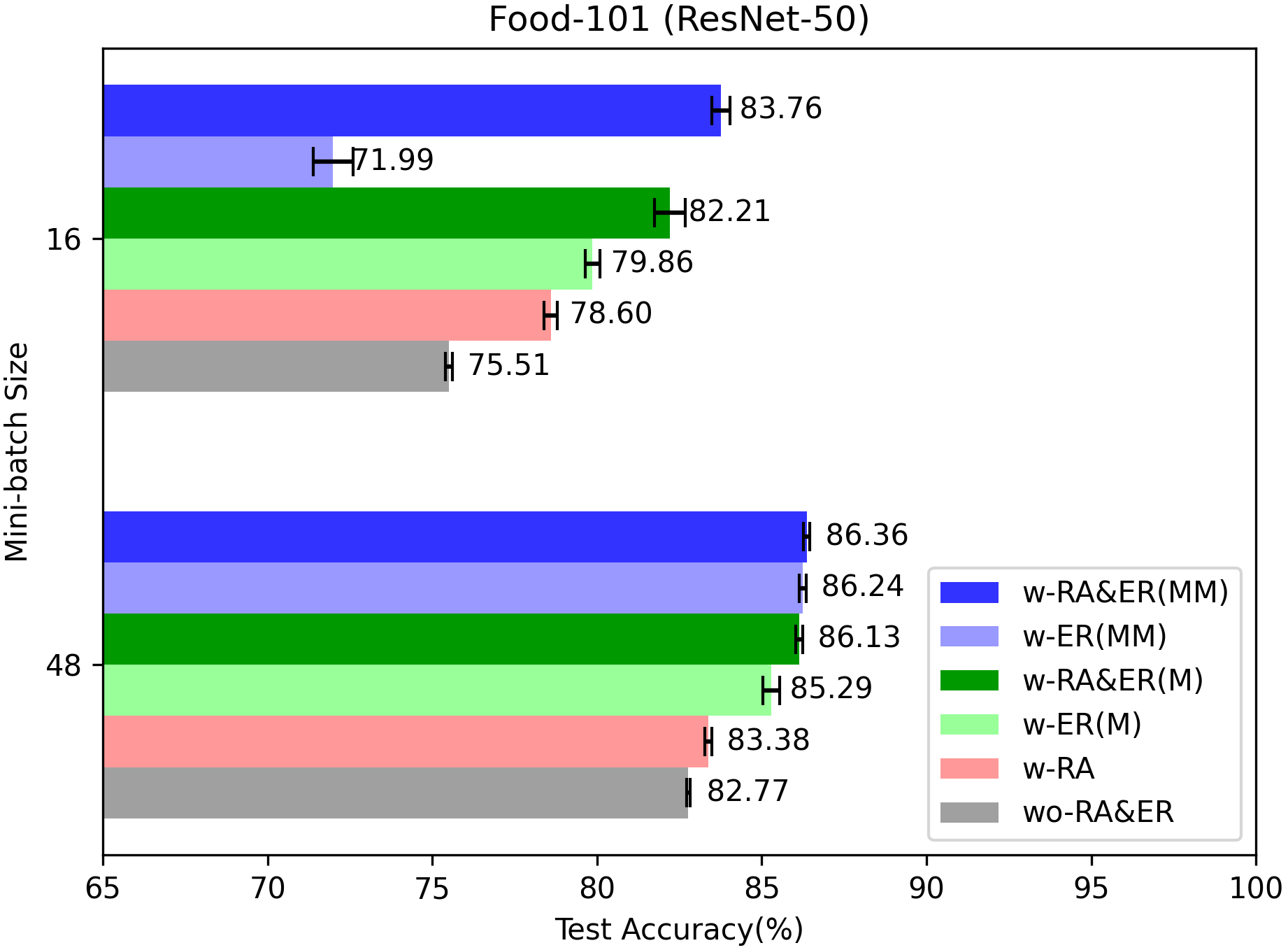}
	\end{minipage}%
 	\begin{minipage}[t]{0.42\linewidth}
		\includegraphics[width=\linewidth]{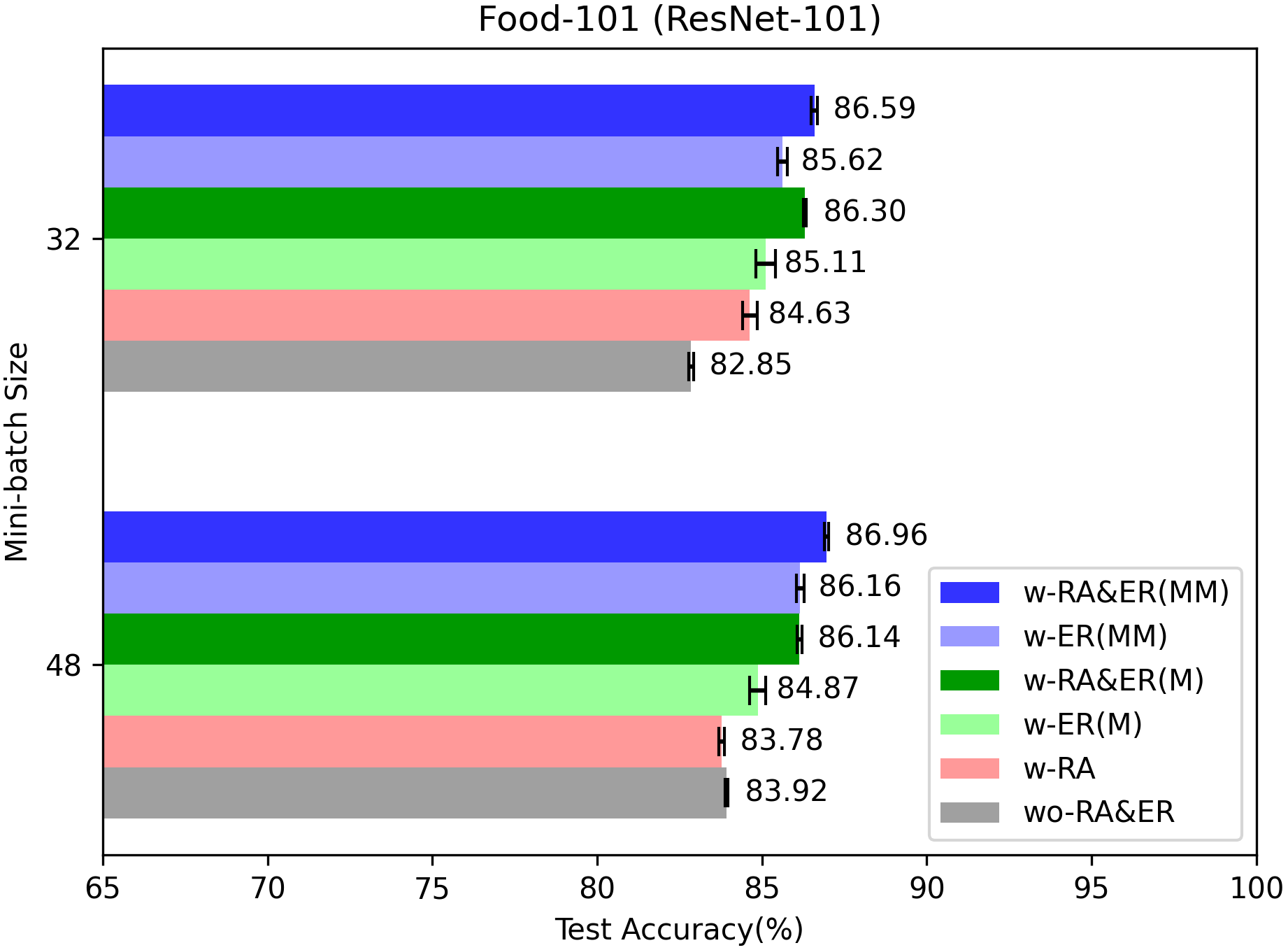}
	\end{minipage}%
 
	\begin{minipage}[t]{0.42\linewidth}
		\includegraphics[width=\linewidth]{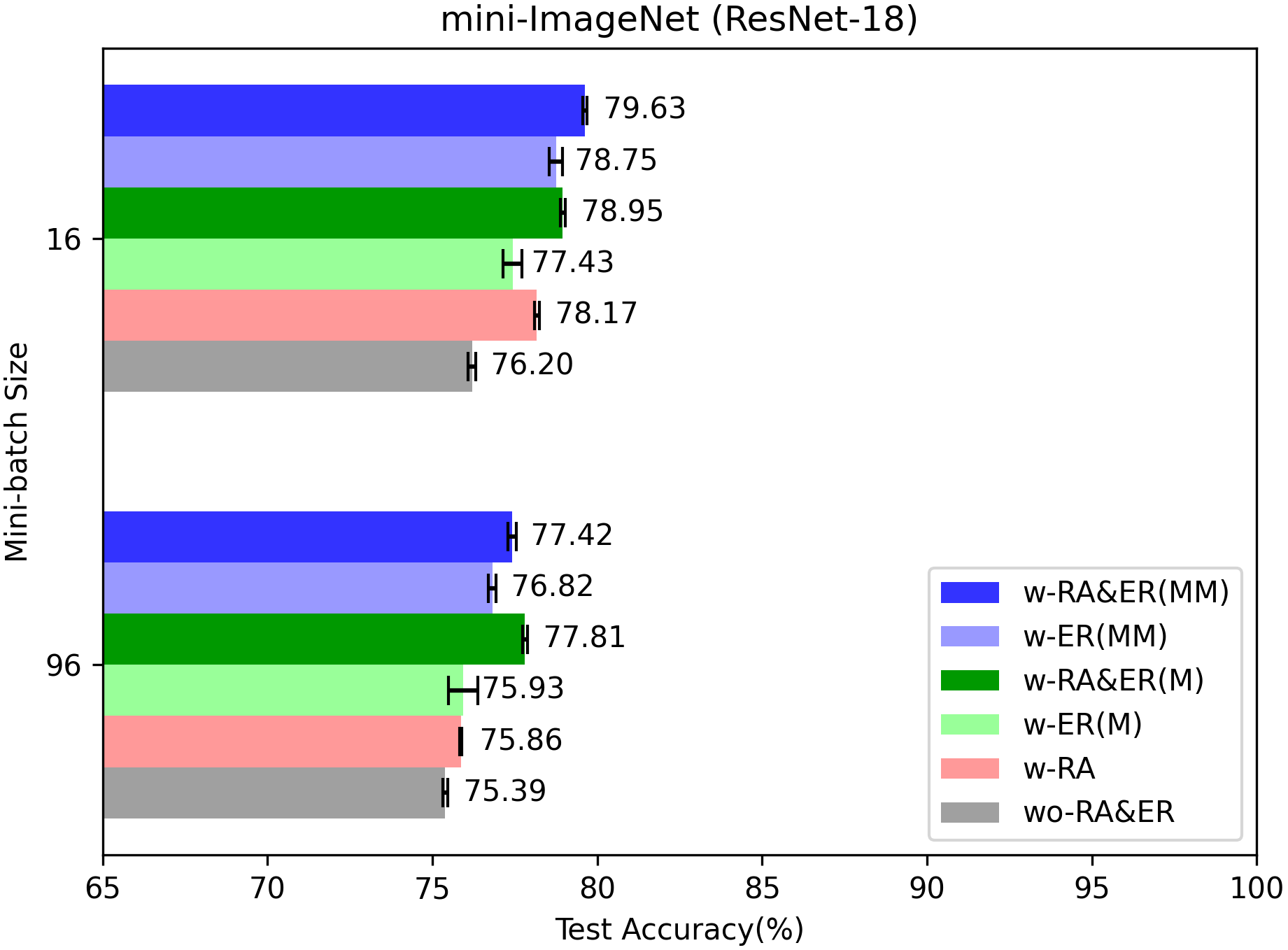}
	\end{minipage}%
	\begin{minipage}[t]{0.42\linewidth}
		\includegraphics[width=\linewidth]{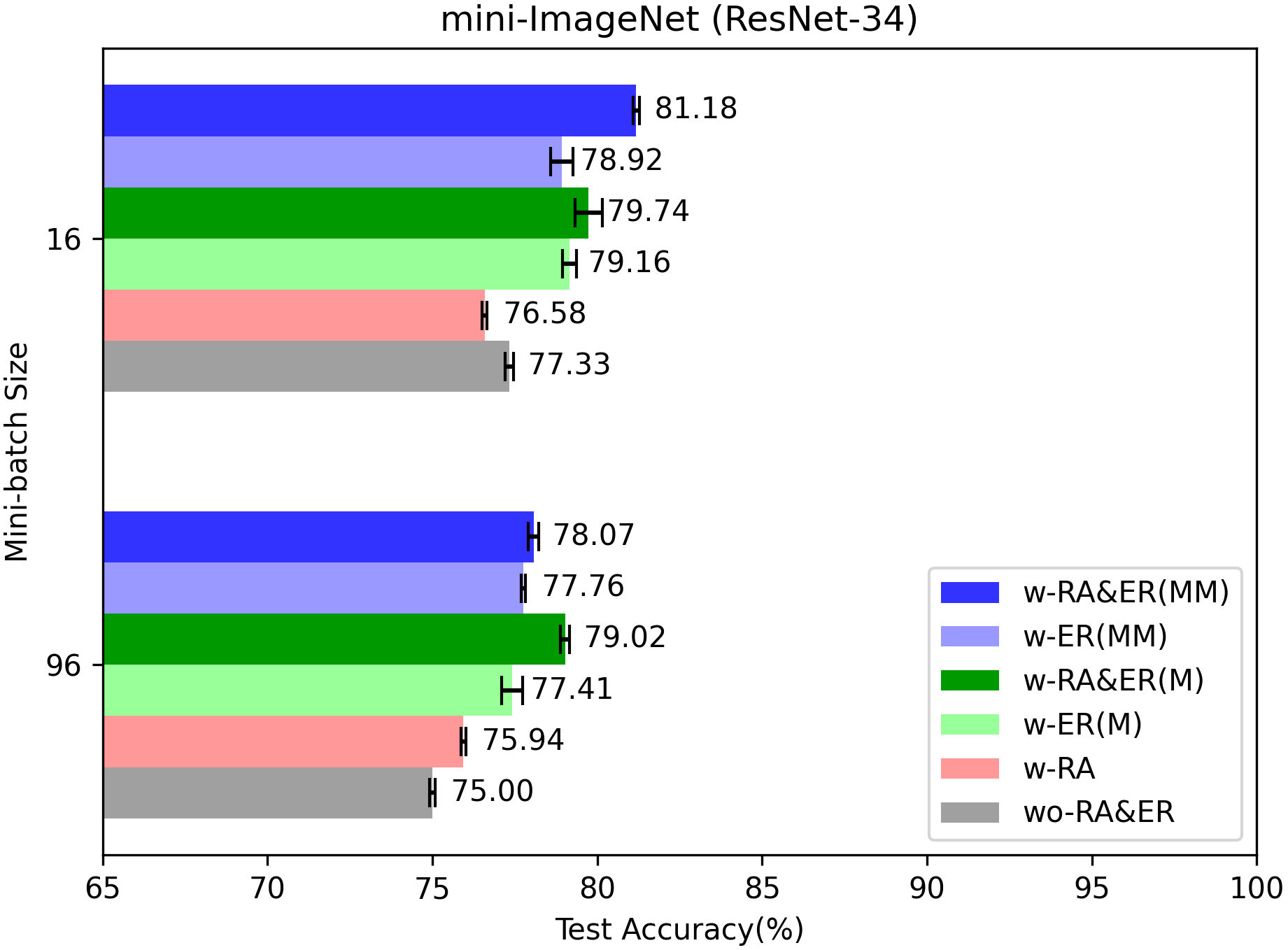}
	\end{minipage}%
	\caption{\label{fig:fig5}Comparison of classification accuracies for \emph{wo-RA\&ER}, \emph{w-RA}, \emph{w-ER(M)}, \emph{w-ER(MM)}, \emph{w-RA\&ER(M)}, and \emph{w-RA\&ER(MM)} across datasets and mini-batch sizes.}
\end{adjustwidth}
\end{figure*}

\subsection{Performance Evaluation: Training from Scratch}
In this subsection, we present and analysis the experimental results of the ResNet models of different depths trained from scratch. Here, our analytical approach begins with an exploration of the performance of \emph{w-RA}, followed by an analysis of \emph{w-ER(M)} and \emph{w-ER(MM)}. Subsequently, we delve into the performance of the synergistic mixup solutions, \emph{w-RA\&ER(M)} and \emph{w-RA\&ER(MM)}, to ascertain their combined effects.
\newline

\begin{figure}[!b]
\begin{adjustwidth}{-2.5cm}{-2.5cm}
\centering
\includegraphics[scale=.55]{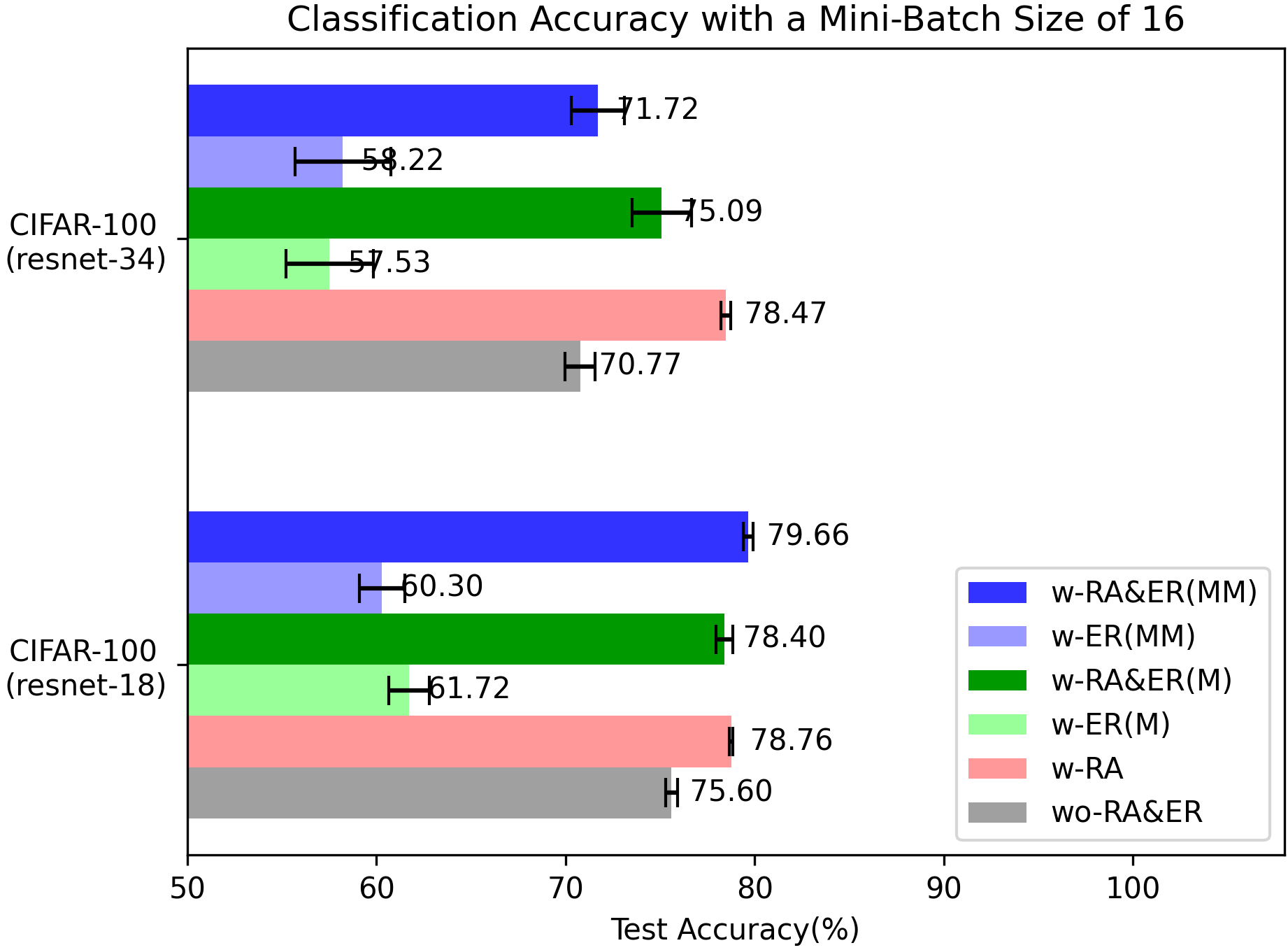}
\caption{\label{fig:fig6}Comparison of classification accuracies for \emph{wo-RA\&ER}, \emph{w-RA}, \emph{w-ER(M)}, \emph{w-ER(MM)}, \emph{w-RA\&ER(M)}, and \emph{w-RA\&ER(MM)} when the mini-batch size is 16.}
\end{adjustwidth}
\end{figure}

\begin{figure*}[!b]
\begin{adjustwidth}{-2.5cm}{-2.5cm}
	\centering
	\begin{minipage}[t]{0.33\linewidth}
		\includegraphics[width=\linewidth]{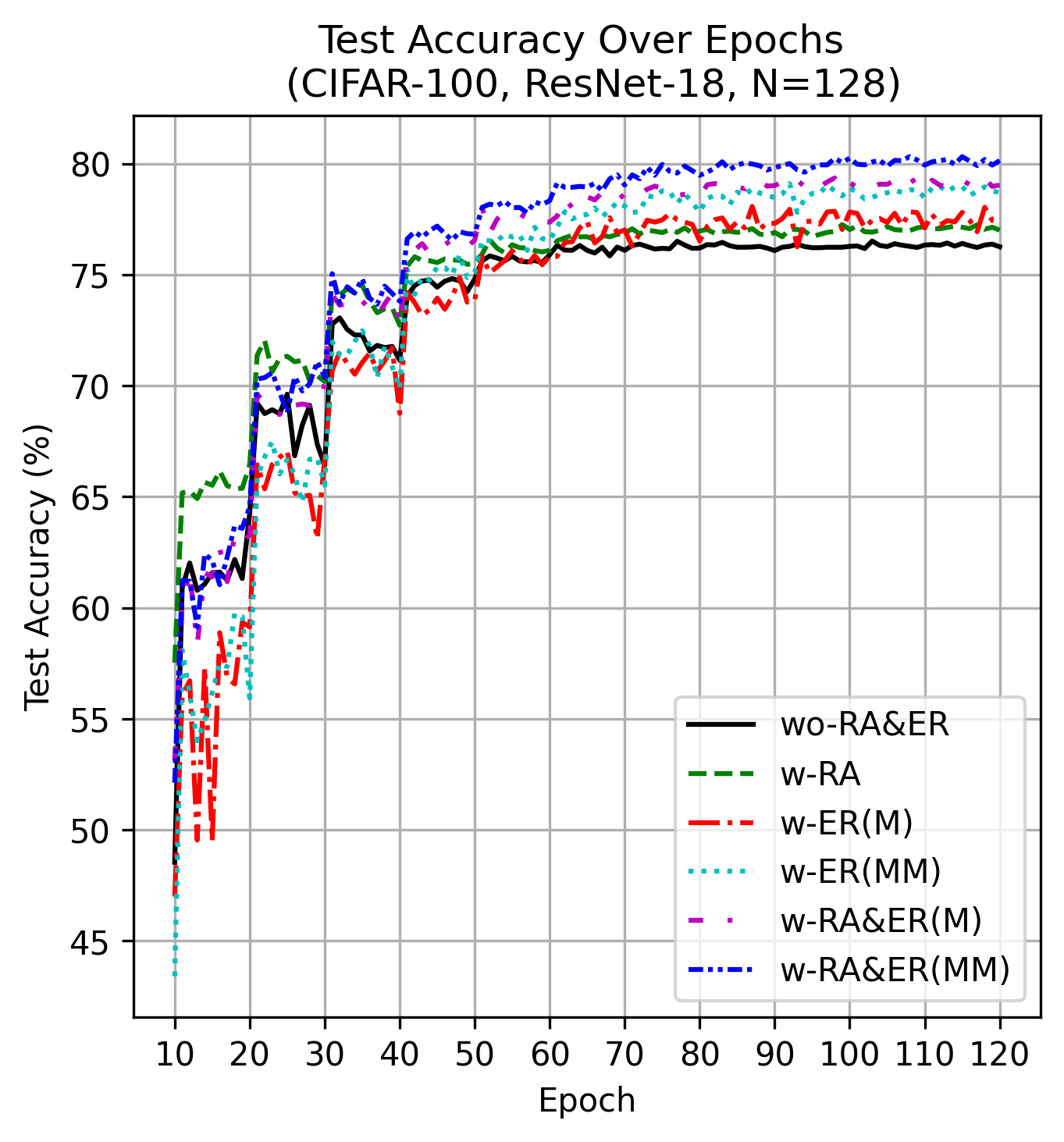}
	\end{minipage}%
	\begin{minipage}[t]{0.33\linewidth}
		\includegraphics[width=\linewidth]{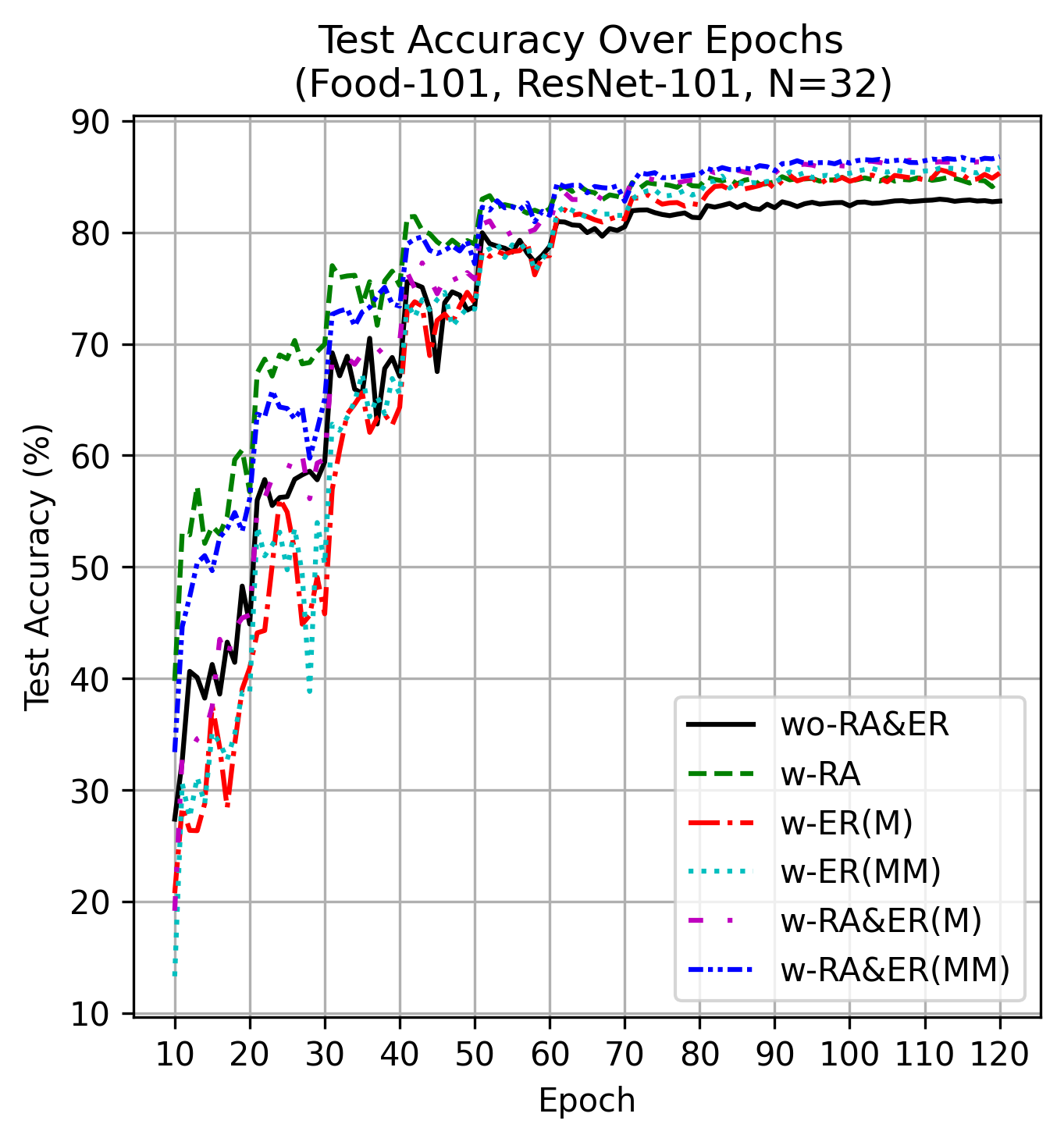}
	\end{minipage}%
	\begin{minipage}[t]{0.33\linewidth}
		\includegraphics[width=\linewidth]{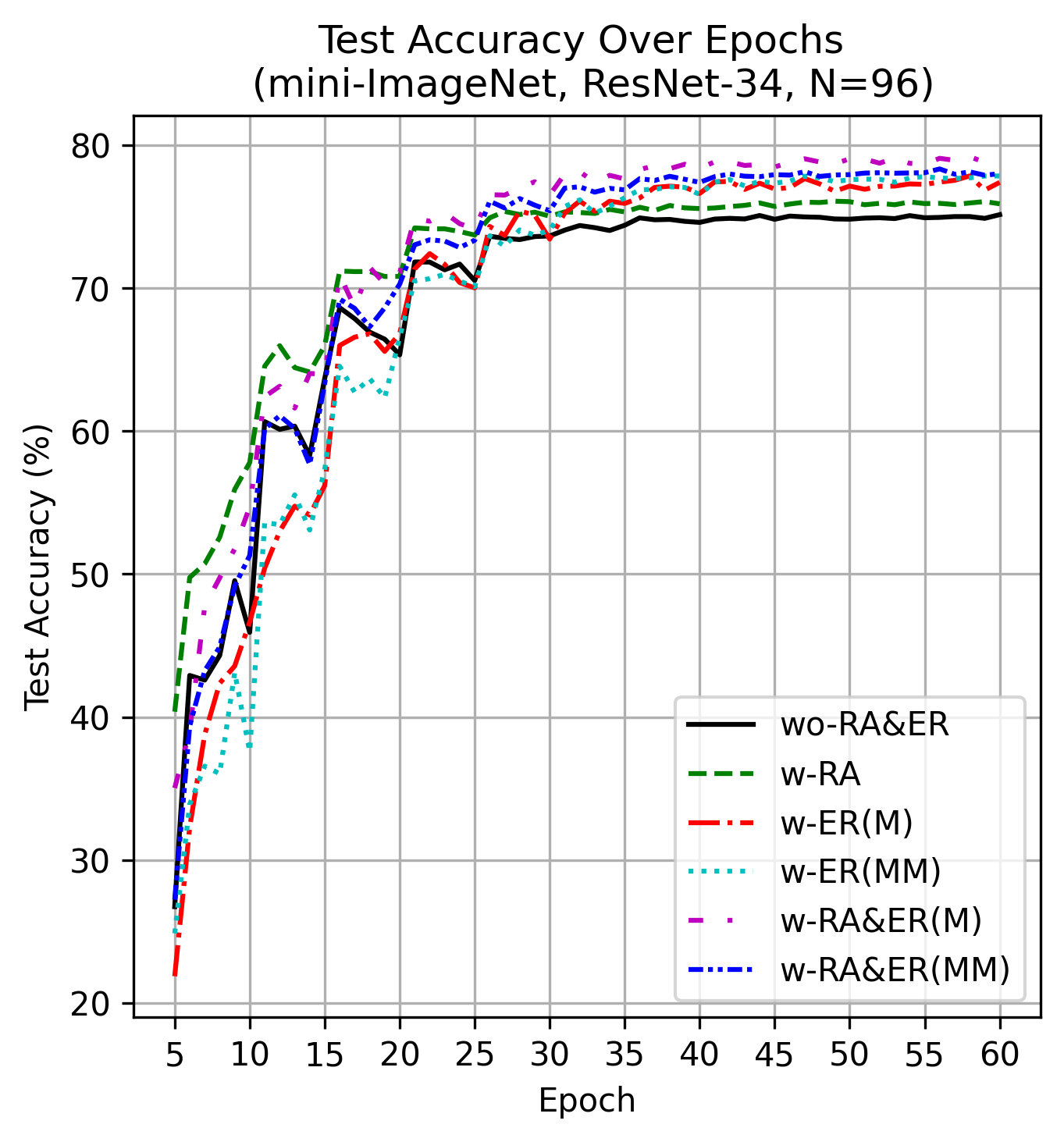}
	\end{minipage}
	\caption{\label{fig:fig7}Comparison of classification accuracies for \emph{wo-RA\&ER}, \emph{w-RA}, \emph{w-ER(M)}, \emph{w-ER(MM)}, \emph{w-RA\&ER(M)}, and \emph{w-RA\&ER(MM)} across epochs on three datasets.}
\end{adjustwidth}
\end{figure*}

\begin{figure*}[!b]
\begin{adjustwidth}{-2.5cm}{-2.5cm}
\centering
\includegraphics[scale=.45]{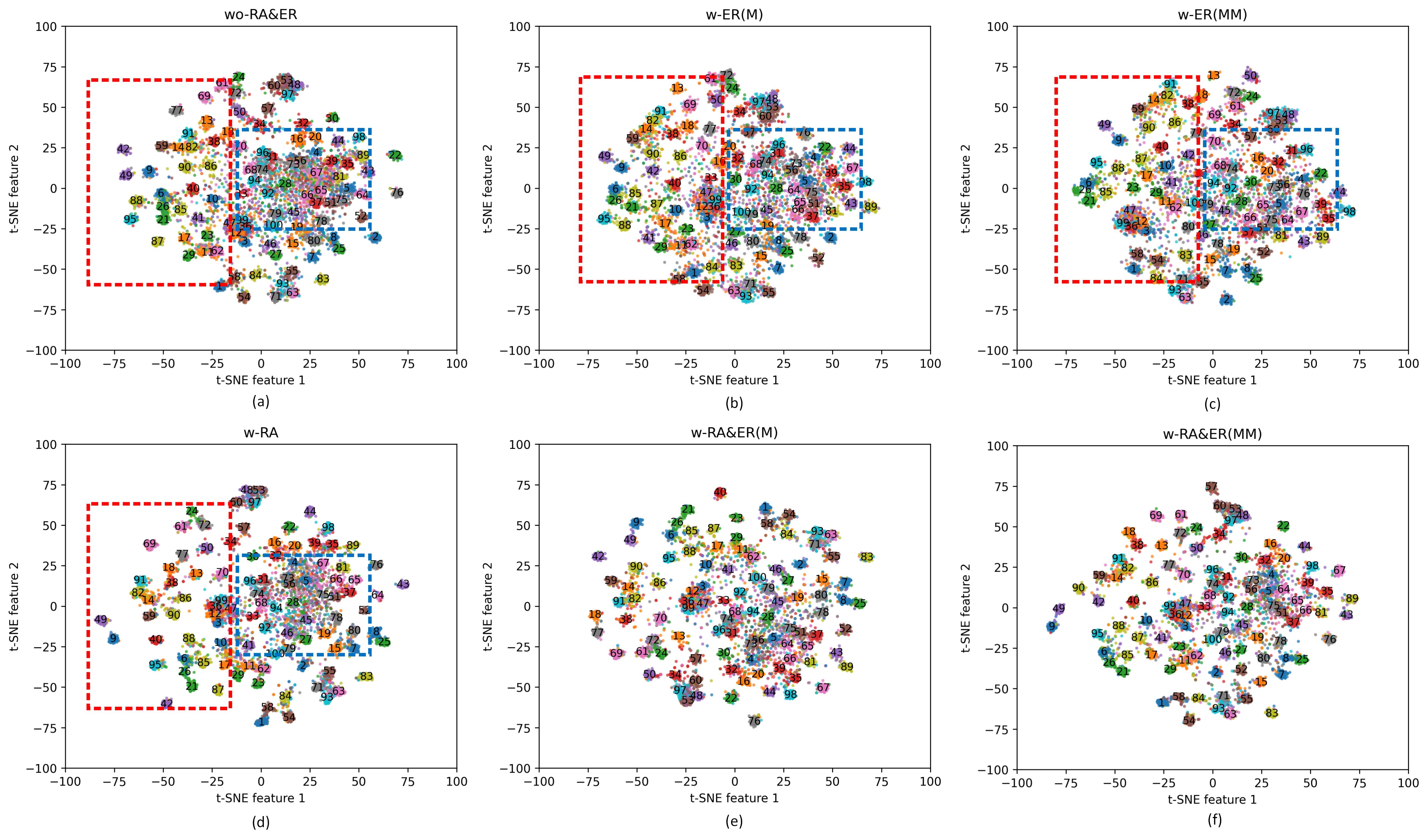}
\caption{\label{fig:fig8}t-SNE visualization of feature representations of testing images from the \emph{CIFAR-100} dataset.}
\end{adjustwidth}
\end{figure*}
\subsubsection{Performance of \emph{w-RA} (\emph{SynerMix-Intra})}
As shown in Figure~\ref{fig:fig5}, an improvement in accuracy was observed across various mini-batch sizes, datasets, and depths of ResNet models when compared to the baseline method, \emph{wo-RA\&ER}. Specifically, \emph{w-RA}, namely \emph{SynerMix-Intra}, outperformed \emph{wo-RA\&ER} in 9 out of 12 experimental scenarios, with accuracy gains ranging from about 0.22\% to 3.09\%, and an average increase of about 1.15\%. In Figure~\ref{fig:fig6}, \emph{w-ER(M)}, \emph{w-ER(MM)}, and \emph{wo-RA\&ER}(model=ResNet-34) exhibited anomalously low classification accuracies due to a lack of stable convergence, whereas \emph{w-RA} remained stable in convergence and achieved high classification accuracies with low standard deviations. These results exhibit that applying the mixing operation to feature representations of unaugmented original images within the same class has the notable potential to improve classification accuracy and robustness to model convergence.

Furthermore,  Figure~\ref{fig:fig7} shows that compared to \emph{wo-RA\&ER}, \emph{w-RA} consistently achieved higher classification accuracy at any given epoch across three datasets. This performance gap is particularly evident in the early training phase. For instance, between the 11th and 20th epochs, \emph{w-RA} exhibited an average accuracy improvement of approximately 14.39\% over \emph{wo-RA\&ER} on the \emph{Food-101} dataset. While the margin of improvement narrows with additional epochs, \emph{w-RA} maintains a consistent lead in accuracy. Besides, it also showed no significant fluctuations, indicating a stable and robust training process. These findings imply that \emph{w-RA} facilitates a rapid convergence of the model in terms of epoch count, likely attributable to more efficient and targeted gradient updates.

We also visualized the distribution of feature representations of testing images using t-SNE plots. As shown in Figure~\ref{fig:fig8}, on a holistic level, the feature representations of each class produced by \emph{w-RA} are more clustered compared to those from \emph{wo-RA\&ER}. This demonstrates that the intra-class mixup in \emph{w-RA} effectively enhances the compactness of the feature space within classes, promoting better discriminability and resulting in improved classification accuracy.

It is worthing note that there are three experimental scenarios where \emph{w-RA} underperforms \emph{wo-RA\&ER} in Figure~\ref{fig:fig5}. This could be attributed to several factors. Firstly, the validation set, comprising a random 10\% subset of the train set, may have a distribution that differs from that of the test set, potentially impacting the selection of the optimal $\beta$ value. This effect was particularly pronounced in the \emph{Food-101} dataset, where the training set contains noisy data and incorrect labels, in contrast to the clean test set \cite{bib6}. Secondly, since there is no guarantee that the synthesized feature representations generated by \emph{w-RA} are sampled from the true distribution, they can mislead the classifier, making it more difficult to correctly distinguish between different classes. This is particularly problematic if the synthesized feature representations of one class overlap with the feature distribution of another class. It increases the likelihood of misclassification and thus reduce the overall accuracy of the model. Notably, this problem can be alleviated by combining with \emph{w-ER}, as illustrated in Section 4.2.3.

\subsubsection{Performance of \emph{w-ER}}
Figure~\ref{fig:fig5} shows that \emph{w-ER(M)} achieved accuracy gains compared to \emph{wo-RA\&ER} in 11 out of 12 experimental scenarios, with improvements ranging from about 0.54\% to 4.35\%, and an average increase of 1.92\%. \emph{w-ER(MM)} also saw accuracy gains in 11 out of 12 experimental scenarios, with increases ranging from about 1\% to 3.47\%, and an average of 2.32\%. Furthermore, it also achieved higher accuracy than \emph{w-ER(M)} in 10 out of the 12 scenarios, by approximately 0.31\% to 1.64\%, averaging 0.97\%. The performance of \emph{w-ER(M)} and \emph{w-ER(MM)} is consistent with the results reported in the literature \cite{bib1, bib2}.

In terms of model convergence, Figure~\ref{fig:fig7} demonstrates that in the early stages of training, \emph{w-ER(M)} and \emph{w-ER(MM)} consistently exhibited lower accuracy than \emph{wo-RA\&ER} and experienced significant fluctuations across multiple epochs with the same learning rate. For example, on the \emph{CIFAR-100} dataset, between the 11th and 20th epochs, \emph{w-ER(M)} were on average 5.71\% less accurate than \emph{wo-RA\&ER}, with the maximum difference between adjacent epochs reaching 9.4\%. However, by the 51th to 60th epoch, the accuracy of \emph{w-ER(M)} had nearly converged to that of \emph{wo-RA\&ER}, and the maximum difference between adjacent epochs was reduced to 0.73\%.

The observed early training instability and slower convergence of \emph{w-ER} can be attributed to the method's inherent randomness. In \emph{w-ER}, training samples are created by randomly blending two images or their hidden representations. This randomness introduces high variability in loss calculations across consecutive mini-batches, increasing the uncertainty of gradient updates. In this case, compared with \emph{wo-RA\&ER} that trains directly using images, there is a higher likelihood of a large deviation between the loss function computed using a mini-batch and the ideal loss function. Consequently, \emph{w-ER} converges more slowly than \emph{wo-RA\&ER}. This issue is exacerbated when the mini-batch size is small, potentially leading to an erratic and unstable convergence path during training. As shown in Figures \ref{fig:fig5} and \ref{fig:fig6}, with a mini-batch size of 16, \emph{w-ER(M)} and \emph{w-ER(MM)} largely underperform compared to \emph{wo-RA\&ER}, with high standard deviations in accuracy.  As training progresses, the model parameters gradually approach the optimum, leading to smaller gradient magnitudes, and the learning rate decreases. Together, these adjustments help to mitigate the negative impact of variability in loss calculations, significantly contributing to more stable convergence. Encouragingly, the initial issue of slow and unstable convergence observed with \emph{w-ER} are mitigated when combined with \emph{w-RA}, as stated in Section 4.2.3.

Additionally, a modest decline in accuracy was observed with the \emph{CIFAR-100} dataset using a ResNet-34 model at a mini-batch size of 32. This decline can be attributed to the fixed setting of the parameter $\alpha$ at 1, which was not finely tuned for different datasets and mini-batch sizes. This setting compromises the models' performance as  \emph{w-ER(M)} and  \emph{w-ER(MM)} are sensitive to the $\alpha$ value. After adjusting $\alpha$ to 0.05, the accuracy of  \emph{w-ER(M)} improved from 78.11\% $\pm$ 0.56\% to 79.71\% $\pm$ 0.40\%, and the accuracy of  \emph{w-RA\&ER(M)} also increased from 82.40\% $\pm$ 0.10\% to 82.52\% $\pm$ 0.10\%. However, when $\alpha$ was set to 2, the accuracy of  \emph{w-ER(M)} dropped to 74.02\% $\pm$ 0.71\%, and the accuracy of  \emph{w-RA\&ER(M)} decreased to 82.12\% $\pm$ 0.08\%.

Delving into the visualization analysis of the feature representation distribution, Figure~\ref{fig:fig8} illustrates that, compared to \emph{wo-RA\&ER}, both \emph{w-ER(M)} and \emph{w-ER(MM)} demonstrate a reduction in the overlap of feature representation distributions across classes, thereby increasing their distinctiveness. This phenomenon is particularly evident in the area enclosed by the blue dashed box in Figure~\ref{fig:fig8}. Upon closer observation, it becomes apparent that \emph{w-ER(MM)} offers a greater degree of alleviation in terms of overlap compared to \emph{w-ER(M)}. For instance, classes 5, 37, 51, 64, 65, 66, and 81 exhibit higher distinctiveness with \emph{w-ER(MM)} than with \emph{w-ER(M)}. This observation supports the results showing that the classification accuracy of \emph{w-ER(MM)} is superior to that of \emph{w-ER(M)}. It is important to note, however, that the feature representation distribution of each class is not tightly clustered, which also impacts the distinctiveness between classes to some extent. This issue can also be effectively mitigated by combining with \emph{w-RA}, as discussed in Section 4.2.3.

\subsubsection{Performance of \emph{w-RA\&ER} (\emph{SynerMix})}
In every experimental setup of Figure~\ref{fig:fig5}, \emph{w-RA\&ER}, namely \emph{SynerMix}, outperformed the individual use of \emph{w-RA} and \emph{w-ER}. Specifically, \emph{w-RA\&ER(M)} achieved an average accuracy increase of 2.61\%, with gains ranging from 0.78\% to 3.95\% over \emph{w-RA}. Against \emph{w-ER(M)}, it improved by an average of 1.66\%, with a range of 0.58\% to 4.29\%. Especially, a synergistic effect was observed in 8 of 12 experimental scenarios such as with the \emph{CIFAR-100} dataset using a ResNet-18 model at a mini-batch size of 32, where \emph{w-RA\&ER(M)} exceeded the sum of the individual gains achieved by \emph{w-RA} and \emph{w-ER(M)} alone. For \emph{w-RA\&ER(MM)}, the improvements were also significant. It surpassed \emph{w-RA} with an average increase of 3.05\%, and gains varied from 1.46\% to 5.16\%. When compared to \emph{w-ER(MM)}, the average increase was 1.97\%, with a range of 0.12\% to 11.77\%. A synergistic effect was also noted in 8 out of the 12 scenarios. Collectively, these findings demonstrate that the benefits (as stated in Section 3.2's third question) of intra-class mixup and inter-class mixup are successfully integrated, significantly enhancing classification accuracy.

It is crucial to note that even when \emph{w-RA} or \emph{w-ER} individually did not surpass \emph{wo-RA\&ER} in accuracy, their combination, \emph{w-RA\&ER}, might still achieve higher accuracy than \emph{wo-RA\&ER}, \emph{w-RA}, and \emph{w-ER}. In the following, we delved into a separate discussion of \emph{w-RA} and \emph{w-ER} to better understand how they complement each other.
\newline

\noindent \textbf{(1) Positive Effects of \emph{w-ER} on \emph{w-RA}}: Within the \emph{CIFAR-100} dataset using a ResNet-34 model, \emph{w-RA} experienced a slight drop in accuracy by 0.41\% compared to \emph{wo-RA\&ER} when the mini-batch size was 128. Despite this initial setback, the combined methods \emph{w-RA\&ER(M)} and \emph{w-RA\&ER(MM)} still demonstrated effectiveness. This pattern was also observed in two other experiments, including with the \emph{Food-101} dataset using a ResNet-101 model at a mini-batch size of 48, and the \emph{mini-ImageNet} dataset using a ResNet-34 model at a mini-batch size of 16. More detailed experimental results related to this phenomenon are presented in Figure~\ref{fig:fig10}. In these three experimental setups, \emph{w-RA} showed instability in accuracy on the validation set across different $\beta$ parameter settings, failing to show any consistent accuracy improvement over \emph{wo-RA\&ER}. However, when combined with \emph{w-ER(M)} or \emph{w-ER(MM)}, there was a stable improvement in accuracy across all $\beta$ parameter settings compared to using \emph{w-ER(M)} or \emph{w-ER(MM)} alone.

While direct experimental evidence is lacking, one plausible explanation for these positive outcomes is that the inter-class mixup in \emph{w-ER} can mitigate the limitations associated with the intra-class mixup in \emph{w-RA}. Specifically, though intra-class mixup in \emph{w-RA} contributes positively to rapid clustering of feature representations within the same class, it also introduces negative effects. As depicted in Section 3.2's second question, one such negative effect is that synthesized feature representations of one class might overlap with the feature distributions of other classes, potentially challenging the classifier's ability to discriminate accurately. However, the inter-class mixup in \emph{w-ER} inherently enhances the distinction among classes, thereby diminishing the chance that synthesized feature representations will be confused with those of other classes. Consequently, the negative impact is mitigated by the integration of \emph{w-RA} with \emph{w-ER}.

\noindent \textbf{(2) Positive Effects of \emph{w-RA} on \emph{w-ER}}: for \emph{w-ER} within the \emph{CIFAR-100} dataset using a ResNet-34 model, \emph{w-ER(M)} underperformed \emph{wo-RA\&ER} in accuracy when the mini-batch size was 32. Nevertheless, \emph{w-RA\&ER(M)} not only closed this gap but also gained an additional 3.65\% in accuracy over \emph{wo-RA\&ER(M)}. The same pattern was observed with \emph{w-ER(MM)} in two other experiments, including with the \emph{Food-101} dataset using a ResNet-50 model at a mini-batch size of 16, and the \emph{CIFAR-100} dataset using a ResNet-18 model at a mini-batch size of 16.

The positive outcomes can be attributed to the intra-class mixup in \emph{w-RA}, which alleviates the limitations of \emph{w-ER}. As illustrated in Section 4.2.2, due to the random blending of images or hidden representations in \emph{w-ER}, loss variability and gradient uncertainty increases. This results in slower and unstable convergence compared to \emph{wo-RA\&ER}. Conversely, the intra-class mixup in \emph{w-RA} largely improves the model's convergence speed and ensures stability, suggesting that it facilitates stable and effective gradient updates. When \emph{w-RA} is integrated with \emph{w-ER}, compared to using \emph{w-ER} alone, the gradient uncertainty is diminished, thereby boosting the model's convergence speed and stability. As shown in Figure~\ref{fig:fig7}, \emph{w-RA\&ER(M)} and \emph{w-RA\&ER(MM)} both demonstrated a significant improvement in accuracy and exhibited greater stability compared to \emph{w-ER(M)} and \emph{w-ER(MM)} during the early stages of training. 

Moreover, \emph{w-RA} indirectly enhances the discriminability between different classes by enhancing the cohesion of feature representations within the same class. This effect occurs because as feature representations of the same class cluster more compactly in the feature space, the model is better able to discern the boundaries that separate different classes. Such a clustering effect diminishes the overlap of feature representations across classes, thereby augmenting the model's capability to distinguish between classes. As shown in Figure~\ref{fig:fig8}, the feature representation distributions for each class generated by \emph{w-RA\&ER(M)} and \emph{w-RA\&ER(MM)} are notably more compact than those produced by \emph{w-ER(M)} and \emph{w-ER(MM)}, thereby enhancing inter-class separability and contributing to a clearer distinction among classes.
\newline

Although \emph{w-RA} and \emph{w-ER} complement each other, their combination \emph{w-RA\&ER} does not guarantee improved accuracy in all scenarios. As shown in Figure~\ref{fig:fig6}, for the ResNet-34 model, \emph{w-RA\&ER(M)} and \emph{w-RA\&ER(MM)} demonstrated notable accuracy gains over \emph{w-ER(M)} and \emph{w-ER(MM)}, but they did not surpass the accuracy of \emph{w-RA} alone. This pattern was also observed with \emph{w-RA\&ER(M)} in the ResNet-18 model. To address this issue, increasing the mini-batch size can enhance the convergence stability of \emph{w-ER}, potentially maximizing the benefits derived from their combination.

\subsection{Performance Evaluation: Fine-Tuning using Pre-Trained Model}

\begin{figure*}[!t]
\begin{adjustwidth}{-2.5cm}{-2.5cm}
	\centering
	\begin{minipage}[t]{0.45\linewidth}
		\includegraphics[width=\linewidth]{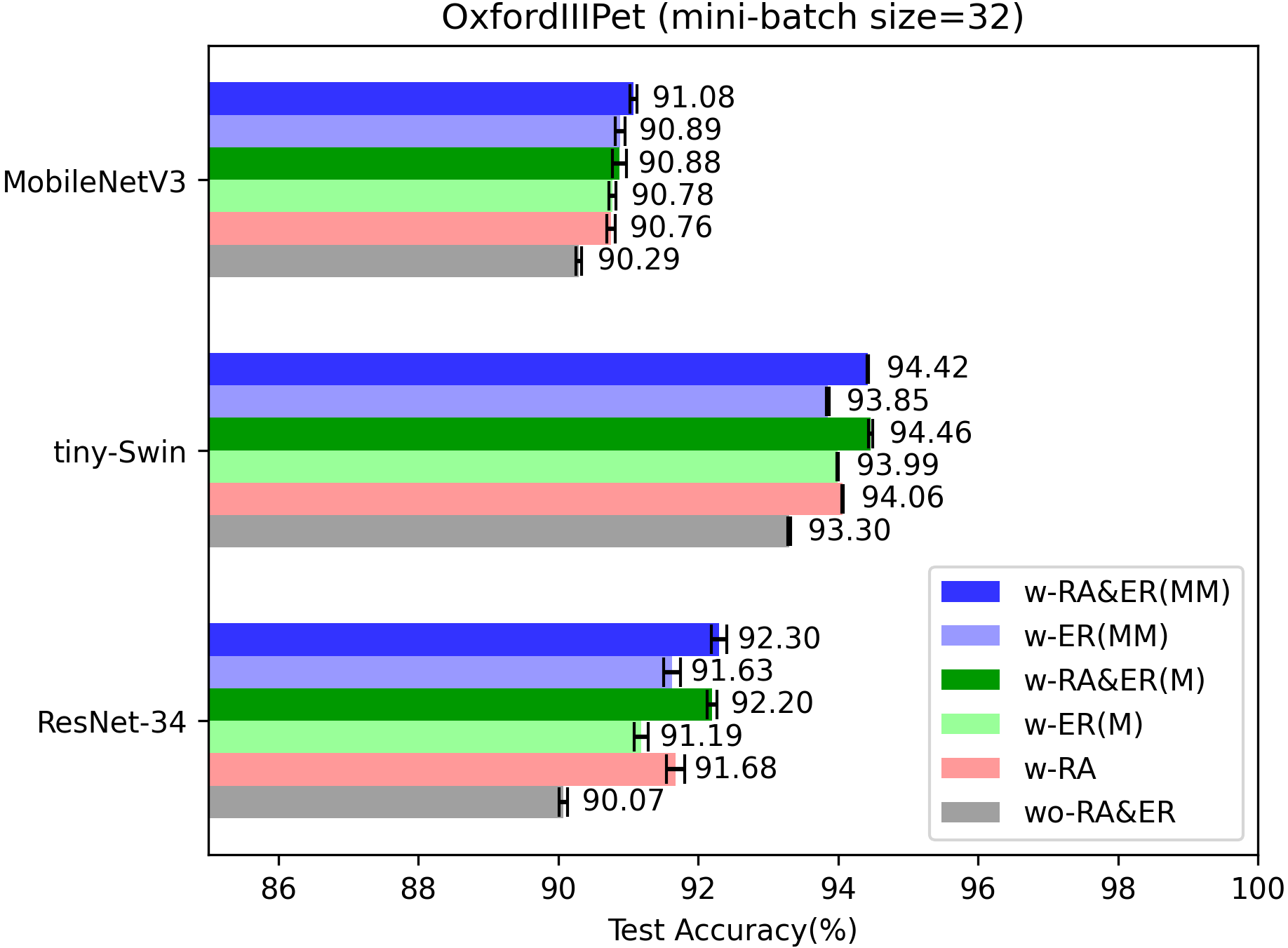}
	\end{minipage}%
	\begin{minipage}[t]{0.45\linewidth}
		\includegraphics[width=\linewidth]{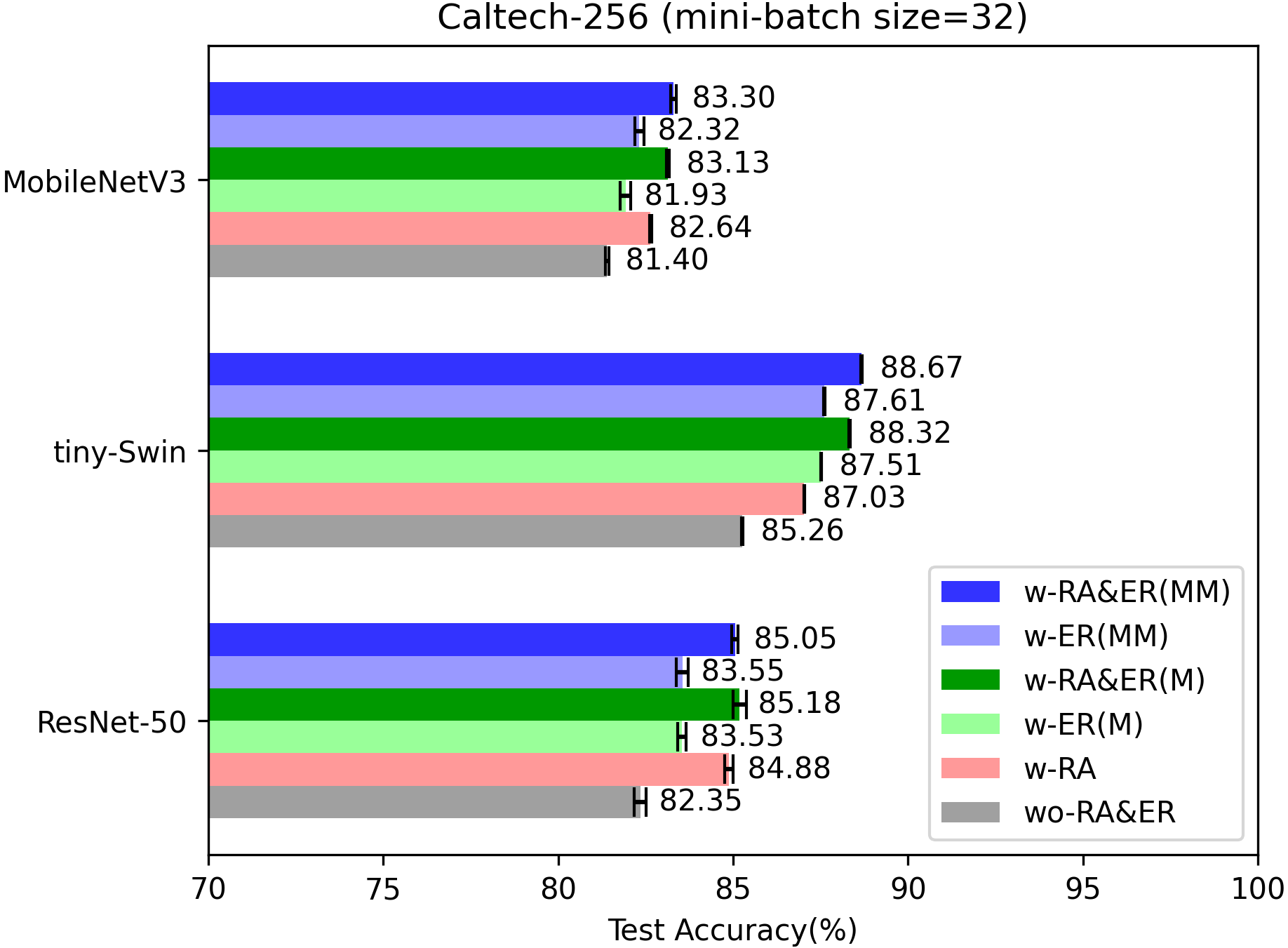}
	\end{minipage}%
	\caption{\label{fig:fig9} Comparison of classification accuracies for \emph{wo-RA\&ER}, \emph{w-RA}, \emph{w-ER(M)}, \emph{w-ER(MM)}, \emph{w-RA\&ER(M)}, and \emph{w-RA\&ER(MM)} when using pre-trained models with different architectures.}
\end{adjustwidth}
\end{figure*}

In this subsection, we report the experimental results of deep learning models with different architectures that were pretrained on the \emph{ImageNet-1K} dataset. As shown in Figure~\ref{fig:fig9}, \emph{w-RA\&ER} still achieved higher classification accuracy than when \emph{w-RA} and \emph{w-ER} were used individually. When these findings are integrated with those from Section 4.2, it becomes evident that \emph{w-RA\&ER}, namely \emph{SynerMix}, is versatile. It proves effective not only for training strategies that begin from scratch but also for fine-tuning pre-trained models. Furthermore, it is adaptable across models of different depths and architectures and is capable of handling image classification tasks with a wide range of classes.

Further analysis revealed that the accuracy gains obtained by \emph{w-RA\&ER} over \emph{wo-RA\&ER} vary among models with different architectures. Specifically, the MobileNetV3-Large model showed the smallest gain, while the tiny-SwinTransformer and ResNet models exhibited more significant improvements. Notably, when the dataset was \emph{OxfordIIIPet}, although the classification accuracy with MobileNetV3-Large using \emph{wo-RA\&ER} was slightly higher than with ResNet-34 by 0.22\%, this advantage disappeared after applying \emph{w-RA\&ER(MM)}, and the accuracy was actually 1.22\% lower. This indicates that the accuracy gains are associated with the model architecture.

Additionally, Figures \ref{fig:fig5} and \ref{fig:fig9} both indicate that higher accuracy with \emph{wo-RA\&ER} is associated with even higher accuracy when using \emph{w-RA\&ER}. This suggests that enhancing the performance of \emph{wo-RA\&ER} is an effective way to improve the accuracy of \emph{w-RA\&ER}. The model architecture is a significant factor influencing the accuracy of \emph{wo-RA\&ER}. Among the three models involved in Figure~\ref{fig:fig9}, MobileNetV3-Large is a lightweight model focused on execution efficiency, with weaker feature extraction capabilities compared to the other two models. In contrast, tiny-SwinTransformer uses a hierarchical transformer structure and a shifted window mechanism to deliver superior feature extraction capabilities. It is evident from Figure~\ref{fig:fig9} that the feature extraction capabilities of these three model architectures are consistent with the experimental results for \emph{w-RA\&ER}. This indicates that the performance of \emph{w-RA\&ER} is influenced by the model architecture.

\begin{figure*}[!p]
\begin{adjustwidth}{-2.5cm}{-2.5cm}
\centering
\includegraphics[scale=.5]{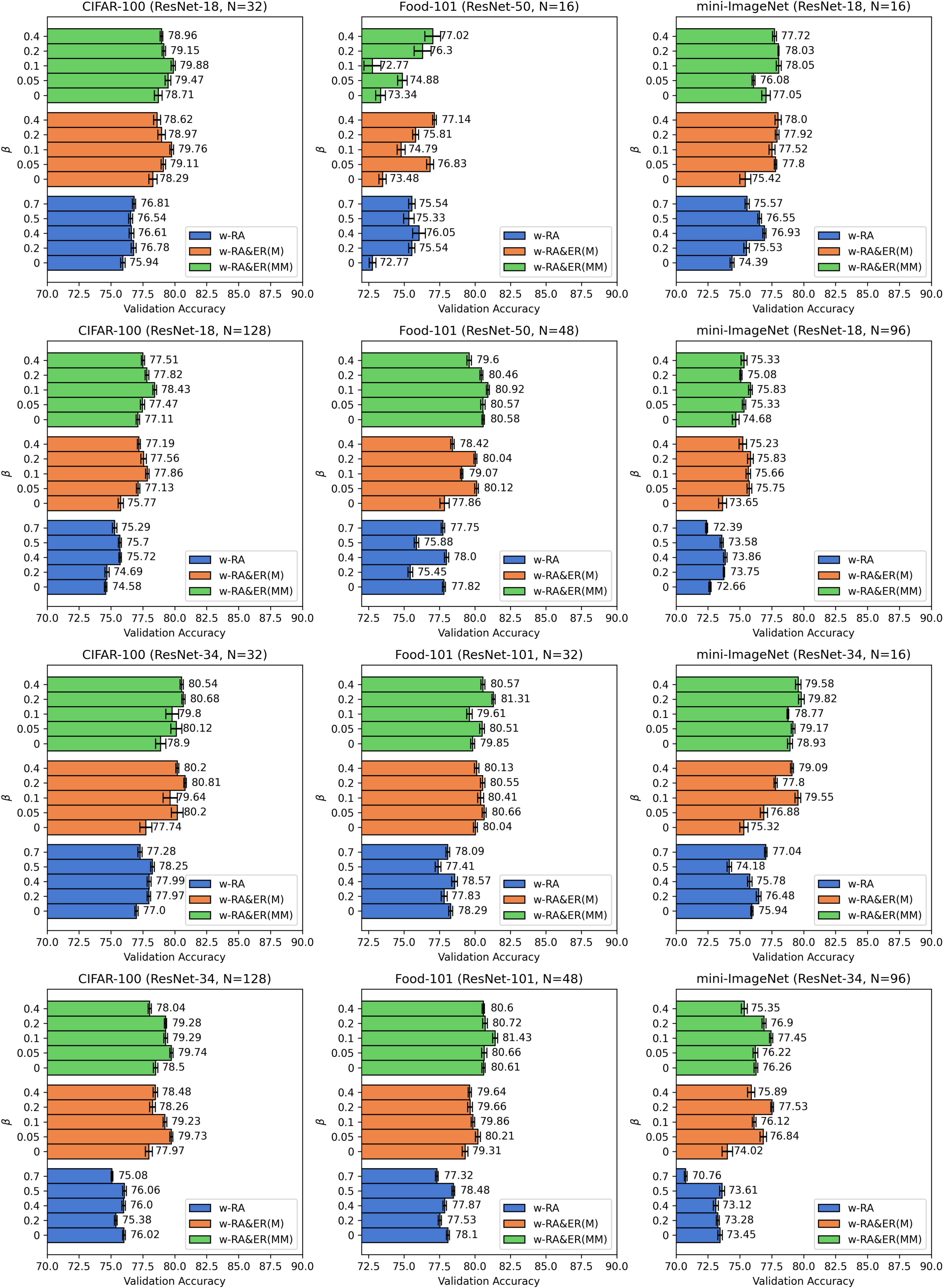}
\caption{\label{fig:fig10}Classification accuracy on the validation set across different conditions with variable $\beta$ values. When $\beta=0$, \emph{w-RA} reduces to \emph{wo-RA\&ER}, \emph{w-RA\&ER(M)} simplifies to \emph{w-ER(M)}, and \emph{w-RA\&ER(MM)} collapses to \emph{w-ER(MM)}.}
\end{adjustwidth}
\end{figure*}

\subsection{Hyperparameter Investigation}
In this subsection, we investigated how the hyperparameter $\beta$ affects the model's classification accuracy on the validation set under various conditions. Subsequently, we assessed the effect of increasing the number $P$ of linear interpolations for each class within a mini-batch—from a single interpolation to multiple. Lastly, we conducted an experimental analysis to ascertain the validity of using feature representations from unaugmented original images, rather than those from augmented images, for generating synthesized feature representations.

\subsubsection{Investigation of the Hyperparameter $\beta$}
$\beta$ is the only new hyperparameter introduced in \emph{SynerMix}, serving to balance the benefits of improving intra-class cohesion against the benefits of augmenting inter-class distinction. The choice of this parameter's value is critical for the model's performance. To explore the impact of this parameter on model performance, we conducted extensive experiments depicted in Figure~\ref{fig:fig10}, which considered variations in model depth, mini-batch size, and dataset.

Figure~\ref{fig:fig10} indicates that the hyperparameter $\beta$ exhibits a stable optimal range for attaining peak accuracy. Specifically, for \emph{w-RA\&ER}, the optimal $\beta$ values consistently fall between 0.05 and 0.2. For \emph{w-RA}, the highest accuracies are achieved with $\beta$ values of 0.4 or 0.5 in 10 out of the 12 experimental scenarios.

In addition, the \emph{Food-101} dataset presents a challenging scenario due to the presence of noise and incorrect labels in the training set, while the test set remains clean. This discrepancy results in the divergent feature distribution between the validation set, which is derived from the training set, and the test set. Despite these challenges, \emph{w-RA\&ER} consistently identifies an optimal $\beta$ that performs well on both the validation and test sets.

Furthermore, \emph{w-RA} does not consistently outperform \emph{wo-RA\&ER} across \{0.2, 0.4\} in 6 out of 12 experimental scenarios. In contrast, \emph{w-RA\&ER(M)} shows a positive gain in accuracy over \emph{w-ER(M)} across \{0.2, 0.4\} in all scenarios. Moreover, it also maintains an accuracy improvement over \emph{w-ER(M)} at two other $\beta$ values of \{0.05, 0.1\}. This suggests that \emph{w-RA\&ER(M)} is highly robust to variations in the hyperparameter $\beta$.

In summary, these findings indicate that \emph{w-RA\&ER} is robust to the hyperparameter $\beta$.

\subsubsection{Investigation of the Number $P$ of Linear Interpolations}
\begin{figure*}[!t]
\begin{adjustwidth}{-2.5cm}{-2.5cm}
	\centering
	\begin{minipage}[t]{0.33\linewidth}
		\includegraphics[width=0.99\linewidth]{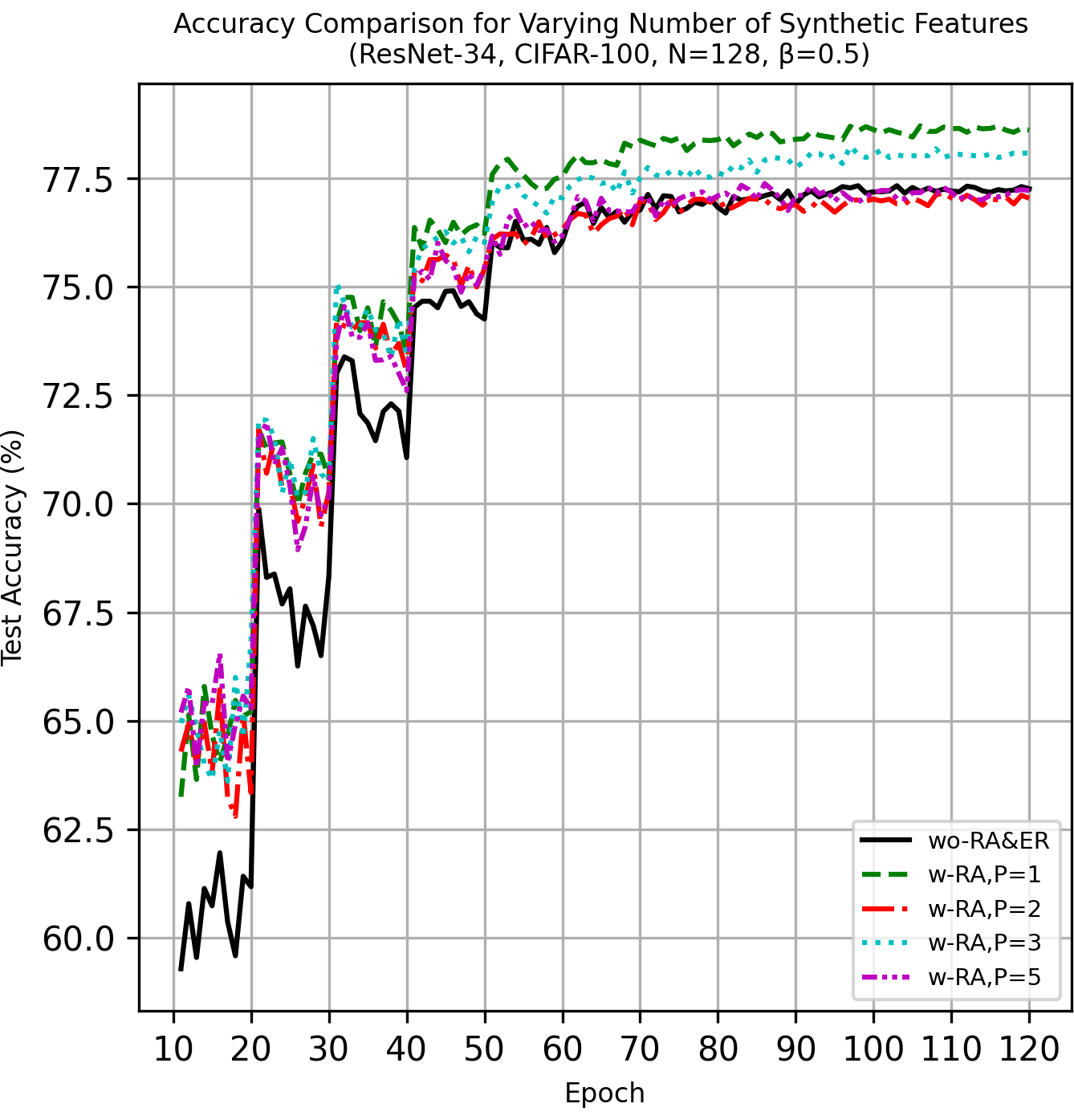}
	\end{minipage}%
	\begin{minipage}[t]{0.33\linewidth}
		\includegraphics[width=0.99\linewidth]{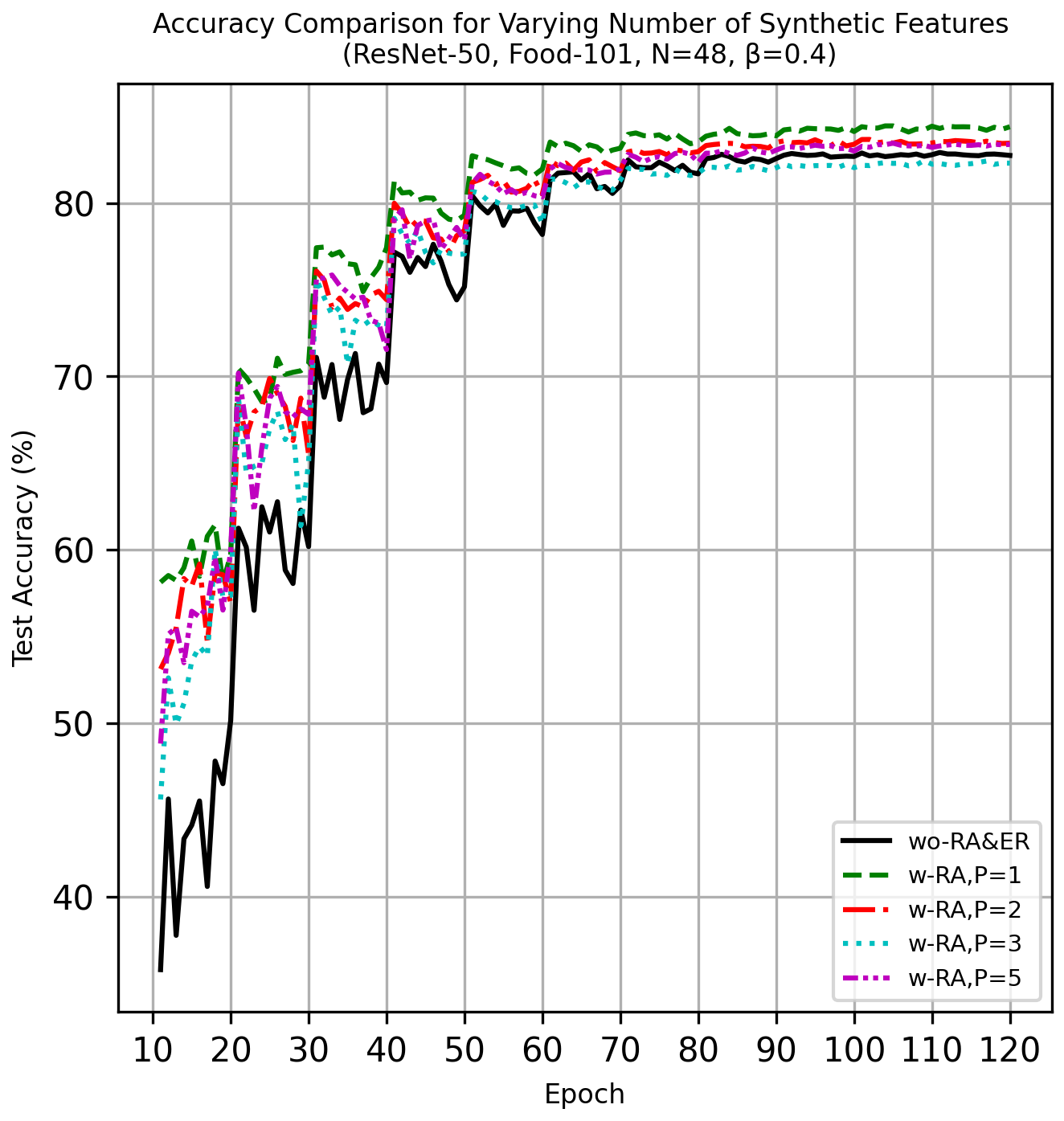}
	\end{minipage}%
	\caption{\label{fig:fig11} Comparison of classification accuracies for \emph{w-RA} with different numbers ($P$=1, 2, 3, 5) of linear interpolations across epochs.}
\end{adjustwidth}
\end{figure*}

In the step of generating synthesized feature representations, we interpolate image feature representations for each class in a mini-batch just once instead of multiple times. Figure~\ref{fig:fig11} provides the rationale for this practice. It shows that increasing the number $P$ of interpolations to 2, 3, or 5 doesn't lead to any improvement in accuracy almost throughout the model's training compared to a single interpolation ($P$=1). This suggests that interpolating once for each class in a mini-batch is enough. 

The reasons for these results are twofold. On one hand, as discussed in Section 3.2's second question, there is no guarantee that synthesized feature representations are sampled from the true distribution. Therefore, an increase in the number of synthesized feature representations could cause the classifier to place less emphasis on the real ones, impairing its ability to classify correctly. On the other hand, as explained in Section 3.2's first question, a synthesized feature representation can be considered a random point within the convex hull or on the line segment formed by the feature representations of the images that generated it. The randomness inherent in the synthesized representations critically contributes to the stochastic nature of the gradients used for optimizing model parameters. When $P=1$, the stochasticity of the gradients is at its peak, and as $P$ increases, this stochasticity progressively diminishes. Since appropriate levels of stochasticity in gradients are known to offer advantages by preventing convergence to local minima, improving model generalization, escaping from saddle points, and promoting exploration during the optimization process, this reduction in stochasticity with $P$ greater than 1 could be another potential reason. 

The mathematical justification for the diminished stochasticity is detailed in the following. Given a mini-batch with $C$ classes, for each class $c$, we have a set of $P$ synthesized feature representations $\bm{f}^c_1, \bm{f}^c_2, \ldots,\bm{f}^c_P$. The loss $\mathscr{L}_{intra}$ is defined as the mean of the softmax losses computed for all synthesized representations across all classes.

The loss $\mathscr{L}_{cp}$ associated with a single feature representation $\bm{f}^c_p$ is characterized by the cross-entropy between the predicted probability distribution and the one-hot encoded true class labels:
\begin{equation}
\mathscr{L}_{cp} = -\sum_{i=1}^{C} \delta_{ic} \log(p_{cpi})
\end{equation}
where $\delta_{ic}$ denotes the Kronecker delta function, which equals 1 if $i = c$ and 0 otherwise. The predicted probability $p_{cpi}$ for class $i$ is articulated by the softmax function:
\begin{equation}
p_{cpi} = \frac{e^{z_{cpi}}}{\sum_{j=1}^{C} e^{z_{cpj}}}
\end{equation}
Here, $z_{cpi}$ represents the logit corresponding to class $i$ for the feature representation $\bm{f}^c_p$.

In the scenario where the parameter is the weight vector \( \bm{w}_i \) within the final fully connected layer associated with class \( i \), given the logit \( z_{cpi} = \bm{w}_i^{\top} \bm{f}_p^c + b_i \), where \( b_i \) represents the bias term corresponding to weight vector \( \bm{w}_i \), the gradient of \( z_{cpi} \) with respect to \( \bm{w}_i \) is given by:
\begin{equation}
\frac{\partial z_{cpi}}{\partial \bm{w}_i} = \bm{f}^c_p
\end{equation}
, and the gradient of the loss $\mathscr{L}_{cp}$ with respect to $\bm{w}_i$ is given by:
\begin{equation}
\frac{\partial \mathscr{L}_{cp}}{\partial \bm{w}_i} = \sum_{i=1}^{C} \left( p_{cpi} - \delta_{ic} \right) \bm{f}^c_p
\end{equation}

To compute the gradient of the aggregate loss $\mathscr{L}_{intra}$ with respect to $\bm{w}_i$, we sum over all classes, feature representations, and predicted class probabilities:
\begin{align}
\frac{\partial \mathscr{L}_{intra}}{\partial \bm{w}_i} &= \frac{1}{C \cdot P} \sum_{c=1}^{C} \sum_{p=1}^{P} \sum_{i=1}^{C} \left( p_{cpi} - \delta_{ic} \right) \bm{f}^c_p \\
&= \frac{1}{C} \sum_{c=1}^{C} \sum_{i=1}^{C} \left(\frac{1}{P} \sum_{p=1}^{P}  \left( p_{cpi} - \delta_{ic} \right) \bm{f}^c_p \right) 
\end{align}

In the scenario where the parameter is the weight $w$ situated in any of the hidden layers, the gradient of the logit $z_{cpi}$ with respect to it necessitates the computation of the derivative of the feature vector with respect to it, which entails additional backpropagation through the network's architecture. In this case, the gradient of the logit $z_{cpi}=\bm{w}_i^{\top}\bm{f}_p^c + b$ with respect to $\bm{f}^c_p$ is:
\begin{equation}
\frac{\partial z_{cpi}}{\partial \bm{f}^c_p} = \bm{w}_{i}^{\top}
\end{equation}
and the gradient of the loss $\mathscr{L}_{intra}$ with respect to the weight $w$ is:
\begin{equation}
\frac{\partial \mathscr{L}_{intra}}{\partial w} = \frac{1}{C \cdot P} \sum_{c=1}^{C} \sum_{p=1}^{P} \sum_{i=1}^{C} \left( p_{cpi} - \delta_{ic} \right) \frac{\partial z_{cpi}}{\partial \bm{f}^c_p} \frac{\partial \bm{f}^c_p}{\partial w}
\end{equation}
Substituting the gradient of the logit $z_{cpi}$ with respect to $\bm{f}^c_p$ yields:
\begin{align}
\frac{\partial \mathscr{L}_{intra}}{\partial w} &= \frac{1}{C \cdot P} \sum_{c=1}^{C} \sum_{p=1}^{P} \sum_{i=1}^{C} \left( p_{cpi} - \delta_{ic} \right) \bm{w}_{i}^{\top} \frac{\partial \bm{f}^c_p}{\partial w} \\
&= \frac{1}{C} \sum_{c=1}^{C} \sum_{i=1}^{C} \bm{w}_{i}^{\top} \frac{ \partial  \left( \frac{1}{P} \sum_{p=1}^{P} \left( p_{cpi} - \delta_{ic} \right)  \bm{f}^c_p \right)}{\partial w}
\end{align}

The stochasticity of the gradients in equations (14) and (18) stems from the randomness of $\bm{f}^c_p$. An examination of the two equations reveals that both include the term$\frac{1}{P} \sum_{p=1}^{P} \left( p_{cpi} - \delta_{ic} \right) \bm{f}^c_p$.With an increasing number $P$,this term converges to its expected value $E\left[ \left(p_{cpi} - \delta_{ic} \right) \bm{f}^c_p \mid \bm{f}^c_p \in \text{the convex hull or line segment of class } c \right]$. Consequently, the stochasticity diminishes. 

\subsubsection{Investigation of Unaugmented Original vs. Augmented Feature Representations}
\begin{figure}[!t]
\begin{adjustwidth}{-2.5cm}{-2.5cm}
\centering
\includegraphics[scale=.5]{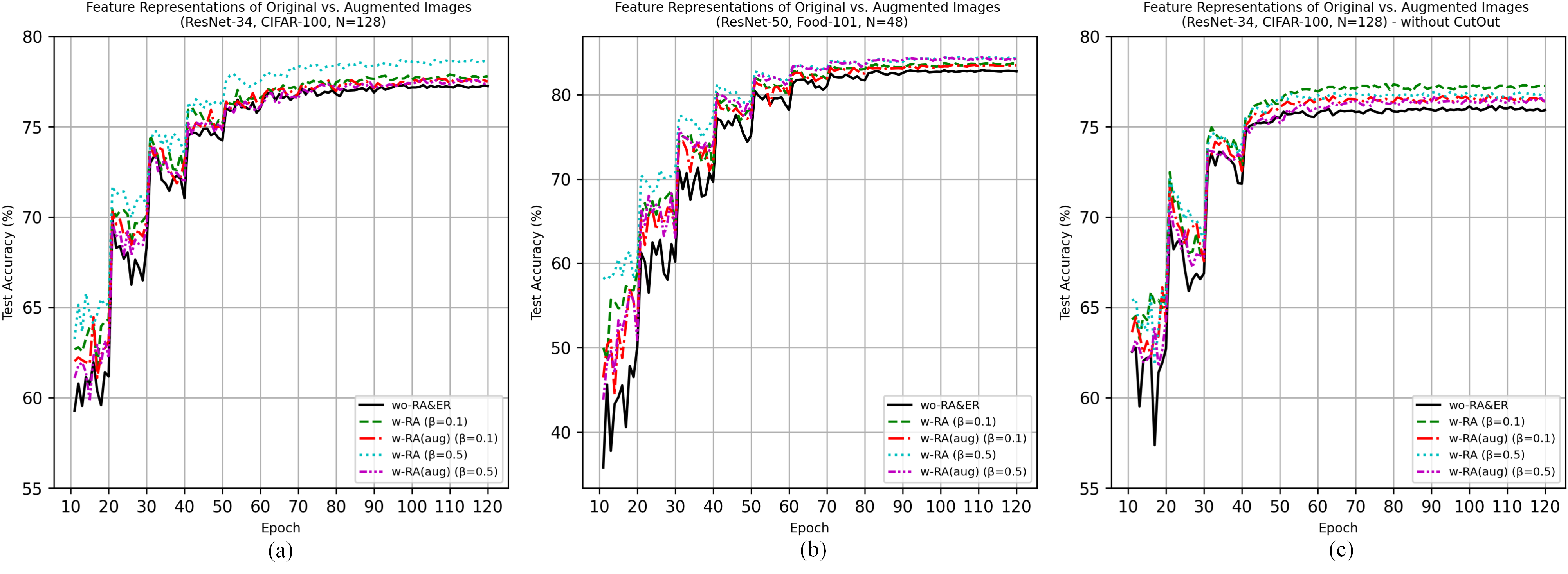}
\caption{\label{fig:fig12}Comparison of classification accuracies of \emph{w-RA} using unaugmented original versus augmented images for generating synthesized feature representations.}
\end{adjustwidth}
\end{figure}

As stated in Section 3.2's second question,  \emph{SynerMix-Intra} prefers the use of feature representations from unaugmented original images over those from augmented images to create synthesized feature representations. The soundness of this practice is corroborated by Figure~\ref{fig:fig12}. Figures \ref{fig:fig12}(a) and \ref{fig:fig12}(b) both illustrate that synthesized feature representations derived from feature representations of unaugmented original images yield superior classification accuracy across two distinct \(\beta\) values compared to those derived from augmented images.

Moreover, with the \emph{CIFAR-100} dataset, the discrepancy in accuracy between using feature representations from unaugmented original images and augmented images becomes more pronounced when \(\beta\) is increased from 0.1 to 0.5. This pattern is not observed with the \emph{Food-101} dataset. This discrepancy can be attributed to the loss of an image's semantic information, which may lead to a deviation of its feature representation from the true distribution. This deviation can further result in a more significant divergence of synthesized feature representations from the true distribution, thereby adversely affecting the model's capability to classify images accurately. The \emph{CIFAR-100} dataset, with its small 32  \(\times\) 32 pixel images that undergo random cropping, flipping, and cutout operations before being input into the model, is particularly susceptible to semantic information loss. Conversely, the \emph{Food-101} dataset, with images rescaled to have a maximum side length of 512 pixels and that only undergo random cropping and flipping, retains more semantic information. As a result, the accuracy gap between using feature representations from unaugmented original versus augmented images for the \emph{Food-101} dataset remains minimal for both \(\beta=0.1\) and \(\beta=0.5\). For the \emph{CIFAR-100} dataset, the increase in the proportion of the gradient from \(L_{intra}\) as \(\beta\) rises from 0.1 to 0.5 amplifies the adverse effects, thus widening the accuracy gap at \(\beta=0.5\). To further substantiate this explanation, the cutout operation was omitted for the \emph{CIFAR-100} dataset, which mitigated the loss of semantic information. In this scenario, as shown in Figure~\ref{fig:fig12}(c), the previously observed phenomenon was no longer evident. This outcome validates the initial explanation. An additional observation is that data augmentation strategies can influence the optimal value of \(\beta\) in \emph{w-RA}. Specifically, a higher accuracy is achieved with \(\beta=0.1\) compared to \(\beta=0.5\) in Figure~\ref{fig:fig12}(c), while the opposite is true in Figure~\ref{fig:fig12}(a), where \(\beta=0.5\) outperforms \(\beta=0.1\).

In conclusion, the above insights justify the use of feature representations derived from unaugmented original images to generate synthesized feature representations.

\subsection{Discussion}
\emph{SynerMix} incorporates both inter-class mixup happened at the input or early hidden layers and intra-class mixup occurred at the penultimate layer before the single fully connected layer dedicated to classification. These layers are present in a variety of model architectures, making \emph{SynerMix} largely agnostic to the specific model architecture. As a result, it can be easily transferred to other domains where current mixup techniques have demonstrated potential, such as in speech and text classification. Additionally, \emph{SynerMix} can be tailored for tasks involving classification components, such as object detection.

Despite the notable gains of \emph{SynerMix} in classification accuracy, it also leads to an almost twofold increase in training costs. This is attributed to \emph{SynerMix-Intra}, which adds a branch that closely mirrors the original model. For example, on the \emph{Food-101} dataset, using a ResNet-101 model with a mini-batch size of 32 on an NVIDIA GTX 4090 graphics card, the epoch duration of \emph{wo-RA\&ER} is roughly 290 seconds. In contrast, the training time of \emph{w-RA\&ER(M)} extends to about 570 seconds per epoch.

Potential avenues for future research include: (1) \emph{SynerMix-Intra} has proven effective in markedly accelerating model convergence, as measured by epoch count. Building upon this, further research could analyze the shifting trajectories of feature representations across epochs. Using these trajectories, predictive synthesized feature representations could be generated to further expedite model convergence, mitigating the increased computational cost problem mentioned above. (2) As illustrated in Section 4.4.2, creating synthesized feature representations within the convex hull or on the line segment introduces beneficial stochasticity into the gradients, potentially contributing to higher accuracy. This observation encourages more in-depth research into the impact of the randomness of synthesized representations within local feature spaces on gradient stochasticity in order to improve classification accuracy.

\section{Conclusion}
Existing mixup approaches often overlook intra-class mixup in terms of mixup strategies (intra-class and inter-class mixup), and fall short in enhancing intra-class cohesion through their mixing operations in terms of mixing outcomes. These limitations constrain their performance in image classification tasks. In response to these shortcomings, this article has undertaken the following work:

(1) We have introduced the novel mixup method \emph{SynerMix-Intra} that leverages feature representations of unaugmented original images from the same class to generate synthesized feature representations, with the aim of bolstering intra-class cohesion. The principle behind the method, along with specific implementation strategies—such as giving preference to feature representations from unaugmented original images over those from augmented images for synthesis, and restricting the generation to one synthesized feature representation per class per batch—has been thoroughly investigated. Our empirical research supports that, compared to the baseline devoid of any mixup operations, \emph{SynerMix-Intra} reduces intra-class variability and boosts classification accuracy by an average of 1.25\% (ranging from 0.22\% to 3.09\%) in 15 out of 18 experimental scenarios. Moreover, it shows considerable promise in markedly accelerating model convergence, as indicated by a reduction in the total number of training epochs required.

(2) Building upon \emph{SynerMix-Intra} and established inter-class mixup methods such as MixUp and Manifold Mixup, we have proposed the synergistic mixup solution \emph{SynerMix}. \emph{SynerMix} incorporates inter- and intra-class mixup in a balanced manner while concurrently enhancing intra-class cohesion and inter-class separability, effectively overcoming the two limitations identified in existing mixup methods. Empirical results from a variety of experimental scenarios indicate that \emph{SynerMix} outperforms the best of either MixUp or \emph{SynerMix-Intra} in accuracy by margins ranging from 0.1\% to 3.43\% (averaging 1.16\%). It also surpasses the superior of either Manifold MixUp or \emph{SynerMix-Intra} by margins ranging from 0.12\% to 5.16\% (averaging 1.11\%).

\emph{SynerMix} is versatile and can be applied to other research areas and tasks, such as speech classification, text classification, and object detection. It also paves the way for new potential research directions, which include: a) developing methods for predictive synthesis of feature representations, which leverage the shifting trajectories of image feature representations across epochs to accelerate model convergence, and b) investigating how the randomness of synthesized feature representations within local feature spaces affects the stochasticity of the gradient, with the aim of enhancing classification accuracy.

\section*{Acknowledgements}
This research was funded by Natural Science Research of Jiangsu Higher Education Institutions of China grant number 22KJD520011, awarded to Ye Xu.

\end{document}